\documentclass[12pt]{article}

\usepackage[margin=1in]{geometry}
\usepackage{times}
\usepackage{microtype}
\usepackage{url}
\usepackage{graphicx}
\usepackage{amsmath,amssymb,bm}
\usepackage{booktabs}
\usepackage{multirow}
\usepackage[dvipsnames]{xcolor}

\makeatletter

\newcommand\ccsdesc[2][]{\ignorespaces}
\newcommand\Description[1]{}           
\newcommand{\keywords}[1]{\par\smallskip\noindent\textbf{Keywords: }#1}
\makeatother

\DeclareRobustCommand{\best}[1]{\textbf{\textcolor{Red}{#1}}}
\DeclareRobustCommand{\second}[1]{\textbf{\textcolor{Blue}{#1}}}
\DeclareRobustCommand{\third}[1]{\textbf{\textcolor{ForestGreen}{#1}}}

\usepackage[numbers,sort&compress]{natbib}
\title{ReNiL: Event-Driven Pedestrian Bayesian Localization Using IMU for Real-World Applications}
\author{Kaixuan Wu, Yuanzhuo Xu, Zejun Zhang, Weiping Zhu,\\ Jian Zhang, Steve Drew, Xiaoguang Niu}  
\date{}                                     

\begin{document}
\maketitle

\begin{abstract}
Pedestrian inertial localization plays a crucial role in mobile and Internet of Things (IoT) services, offering ubiquitous positioning without requiring external infrastructure. However, existing deep-learning-based pedestrian inertial localization methods rely on fixed sliding-window integration. These methods lack flexibility in adapting to varying motion scales and cadences, making them difficult to transfer across different devices, users, and motion patterns. Additionally, they cannot estimate uncertainties and errors consistent with the data source. This limits their utility in real-world mobile scenarios. In this paper, we propose Relative Neural Inertial Locator (ReNiL), a novel deep learning-based event-driven Bayesian inference framework for accurate, efficient, and uncertainty-aware pedestrian localization. ReNiL introduces the concept of Inertial Positioning Demand Points (IPDPs) and focuses estimation on contextually meaningful points rather than continuous dense tracking. It supports inference on inertial measurement unit (IMU) sequences at any scale, enabling flexible positioning cadences aligned with application needs, thus saving significant computational overhead. ReNiL integrates a motion-aware orientation filter with an Any-Scale Laplace Estimator (ASLE), a dual-task deep network that combines patch-based self-supervision and Bayesian regression. By parameterizing displacement with a Laplace distribution, ReNiL yields a high-precision estimate of the displacement as well as its Euclidean-homogeneous uncertainty estimate. Then a Bayesian inference chain connects successive IPDPs and other positioning sources to form an efficient and unified localization process. Extensive experiments on RoNIN-ds and our new WUDataset, which feature complex indoor/outdoor motion from 28 participants using multiple different types of devices, have demonstrated that ReNiL achieves state-of-the-art position estimation accuracy and uncertainty consistency, outperforms TLIO, CTIN, iMoT, and RoNIN variants. Application studies have confirmed ReNiL’s ubiquity, robustness and usability for mobile and IoT localization tasks, establishing it as a scalable, uncertainty-aware foundation for next-generation mobile positioning systems.
\end{abstract}



\section{Introduction}

The ubiquitous use of smart mobile devices has increased the demand for location-based services. Pedestrian localization is an important technique that supports indoor navigation, AR/VR interaction, and safety-critical guidance. 
The goal of a pedestrian localization system is to deliver accurate position estimates together with well calibrated error characterization at the time decisions are required. 
Existing approaches fall into two classes: (i) \emph{infrastructure- or map-dependent} methods, including radio fingerprinting and ranging with WiFi, BLE, UWB, or RFID~\cite{shang2022overview,kriz2016improving,gezici2009position,bahl2000radar,xu2023principle,10.1145/3699792,10.1145/3678554,10.1145/3214287} and vision- or LiDAR-based odometry (VIO/LIO), which typically requires prebuilt maps or rich visual texture~\cite{scaramuzza2011visual,zhang2014loam,10.1145/3161409,10.1145/3757940.3757942,10.1145/3669721.3669749}; (ii) \emph{inertial} methods that operate without external signals. Inertial methods use an inertial navigation system (INS) to integrate a specific force and angular rate measured by an inertial measurement unit (IMU). 

This process has gain traction in recent years as the IMU propagates the state over time without infrastructure or prior maps, improving privacy, and simplifies deployment.
In addition, the adoption of deep learning for consumer-grade MEMS IMUs has further improved IMU-only localization. Compared to classical pedestrian dead-reckoning (PDR) and strapdown INS with ZUPT or magnetometer updates~\cite{titterton2004strapdown,anguita2013public}, data-driven methods show that convolutional-, recurrent-, and attention-based models infer relative displacement from IMU sequences and surpass traditional smartphone baselines~\cite{herath2020ronin,liu2020tlio,wald2019rio,8937008,sun2021idol,rao2022ctin,nguyen2025imot,10.1145/3534594}. Probabilistic models further estimate heteroscedastic uncertainty. These models are typically under a Gaussian assumption and trained by minimizing the negative log-likelihood, which facilitates downstream fusion~\cite{liu2020tlio,sun2021idol,8937008,rao2022ctin}. Moreover, IMU can be combined with external sources, such as GNSS, radio, vision, or LiDAR using filtering or factor-graph formulations, for higher accuracy ~\cite{al2021map,magsi2024comparison,lee2021indoor,thio2021fusing,10.1145/3722570.3726876,10.1145/3662009.3662018,10.1145/3214270,10.1145/3328918,10.1145/3569467}.

While great progress has been made facilitating higher performance of pedestrian localization, most deep-learning based inertial methods use a fixed sliding window and directly regress an IMU segment to the displacement or velocity of uniformly spaced trajectory points. This design generates acceptable performance in offline evaluations, but faces three challenges in real-world, online deployment:

\begin{itemize}
\item \textbf{Cadence mismatch and compute waste.} Real-world applications are event driven (e.g., doorway crossings, turns, stairs transitions, arrivals at points of interest). These events are not triggered in fixed time windows. Fixed-rate inferences are excessive and creates unnecessarily dense samples that accumulate drift. The inference cadence does not match decision times, and increases device resource usage.
\item \textbf{Uncertainty miscalibration for fusion.} Many networks either omit uncertainty or assume a single Gaussian model. Nonstationary motion and sensor jitter, where heavy-tailed errors are common. Empirical coverage and negative log-likelihood diverge from the residual statistics. As a result, outputs lack a reliable scale in a shared Euclidean state space, which hinders principled fusion with other sources and degrades subsequent filtering and smoothing.
\item \textbf{Single-scale bias and poor transfer.} A fixed window cannot capture both short, high-dynamic segments and long, slowly varying context. The resulting scale mismatch reduces transfer across users, carry patterns, and devices, and yields brittle performance in heterogeneous environments.
\end{itemize}

To address the challenges above, we propose \textbf{ReNiL}, a deep learning-based Bayesian inference framework for inertial localization.
To eliminate excessive estimation of relative displacement and uncertainty, we define an innovative concept called \textit{inertial positioning demand points (IPDPs)}. ReNiL only compute the estimates at IPDPs.
ReNiL also incorporates a motion-aware orientation filter specifically designed for pedestrian dynamics, enhancing the alignment between IMU measurements and the navigation coordinate system. To support IMU sequences of any scale and enable uncertainty estimation, it employs a dual-task deep network, \textit{Any-Scale Laplace Estimator (ASLE)}, which combines patch-wise self-supervision with Bayesian regression. Furthermore, ReNiL establishes a principled Bayesian inference pipeline that forms a continuous chain of displacement and uncertainty estimates across IPDPs.

When the user starts using the inertial service, the system collects IMU data to adaptively estimate the device orientation and gradually align the navigation coordinate system with the ENU coordinate system. ReNiL first aligns the entire IMU sequence with the navigation coordinate system based on the device orientation at each timestamp and then patches the data according to a given patch length. The patched data are processed in parallel by an input module and a feature extractor to extract motion features. To improve temporal correlation, the features are reconstructed into robust contextual motion features by a contextual builder. A scale-pooling multi-head self-attention mechanism is designed to build unified global features and contextual features, allowing it to regress with different scales. The self-supervision patch method is used to further improve the noise resistance of the model and the contextual learning capabilities. To handle potential aleatoric uncertainty in the data and improve the numerical stability of regression, displacement and its uncertainty estimates are parameterized using the Laplace distribution~\cite{laplace1774memoire}. Additionally, ReNiL provides a general process based on Bayesian inference for pure IPDP tasks or tasks loosely coupled with external positioning sources.

Extensive experiments on the open-source RoNIN-ds and our own WUDataset have demonstrated that ReNiL achieves state-of-the-art position estimation accuracy and uncertainty consistency, surpassing all advanced models. We have verified the validity of parameterizing the displacements with the Laplace distribution and validated its Euclidean homogeneity. We have analyzed the computational overhead of ReNiL and several other models, and verified that ReNiL can greatly reduce the parameters and overhead of floating-point operations. Finally, we have conducted a series of practical application experiments, demonstrating the high availability of ReNiL in real-world scenarios by combining the Bayesian process.

In summary, we make the following contributions.
\begin{itemize}
    \item \textbf{IPDPs task formulation.} To the best of our knowledge, this is the first paper that defines pedestrian localization as an on-demand task with IPDPs, instead of using fixed-rate estimation.
    \item \textbf{ReNiL framework.} We develop ReNiL, an end-to-end deep Bayesian inertial localization system that (i) integrates a motion-aware orientation filter for robust pedestrian orientation alignment; (ii) introduces the ASLE to jointly regress Laplace parametrized displacement and uncertainty from IMU sequences of any scale; and (iii) designs an Euclidean homogeneous Bayesian Inference chain, that propagates uncertainties across successive IPDPs and facilitates loose fusion with external positioning sources.
    \item \textbf{Comprehensive validation.} Experiments on RoNIN-ds and our WUDataset show that ReNiL achieves state-of-the-art position estimation accuracy, Euclidean homogeneous uncertainty, and markedly reduces FLOPs.
\end{itemize}

\section{Related Work}
\label{sec:related-work}

\subsection{Pedestrian Dead Reckoning}
Pedestrian Dead-Reckoning (PDR) estimates a pedestrian’s position by integrating step detection, stride length estimation, and heading direction from inertial sensors~\cite{brajdic2013walk}. Traditional PDR methods are highly dependent on heuristic models and sensor calibration, which may suffer from cumulative drift and noise. Recent work has focused on improving robustness by using map information~\cite{kang2021indoor} or combining additional signals such as magnetometers and barometers~\cite{liu2024reliable}. However, conventional PDR still faces challenges in complex indoor environments with magnetic disturbances and sensor biases~\cite{hou2020pedestrian}.

\subsection{Deep Learning for Inertial Navigation}
Deep learning has been increasingly applied to inertial navigation to address the limitations of traditional algorithms. Data-driven methods learn representations from raw or aligned IMU data to directly estimate displacement. RNNs, especially LSTM networks, have shown effectiveness in modeling temporal dependencies in IMU sequences~\cite{8937008,sun2021idol}. CNN-based architectures have been explored to effectively suppress short-term drift in sensor data~\cite{herath2020ronin,liu2020tlio,wald2019rio}. Attention-based architectures have also been explored to capture long-range correlations in sensor data~\cite{rao2022ctin,nguyen2025imot}. These methods reduce the need for manual calibration and demonstrate improved accuracy.

\subsection{Uncertainty Estimation}
In the context of deep learning, uncertainty modeling has emerged as a critical technique to improve model reliability and robustness, particularly in domains such as computer vision and trajectory prediction.~\cite{gal2016dropout} introduces the concept of dropout training as an approximate Bayesian inference method for deep Gaussian processes, providing a robust tool for modeling model uncertainty.~\cite{kendall2017uncertainties} distinguishes between two primary types of uncertainty: data uncertainty (aleatoric uncertainty) and model uncertainty (epistemic uncertainty). They have proposed a Bayesian deep learning framework that integrates these uncertainties, which has been applied to tasks such as semantic segmentation and deep regression. HiVT~\cite{zhou2022hivt} is proposed to enhance the accuracy and efficiency of motion prediction by modeling interactions between multiple agents through a hierarchical structure.~\cite{gu2024producing} extends various online map estimation methods by incorporating uncertainty estimation and strengthening the role of map information in trajectory prediction.

\section{System Overview and Formulation}
\label{sec:formulation}

\begin{figure*}[!t]
  \centering
  \includegraphics[width=0.95\textwidth]{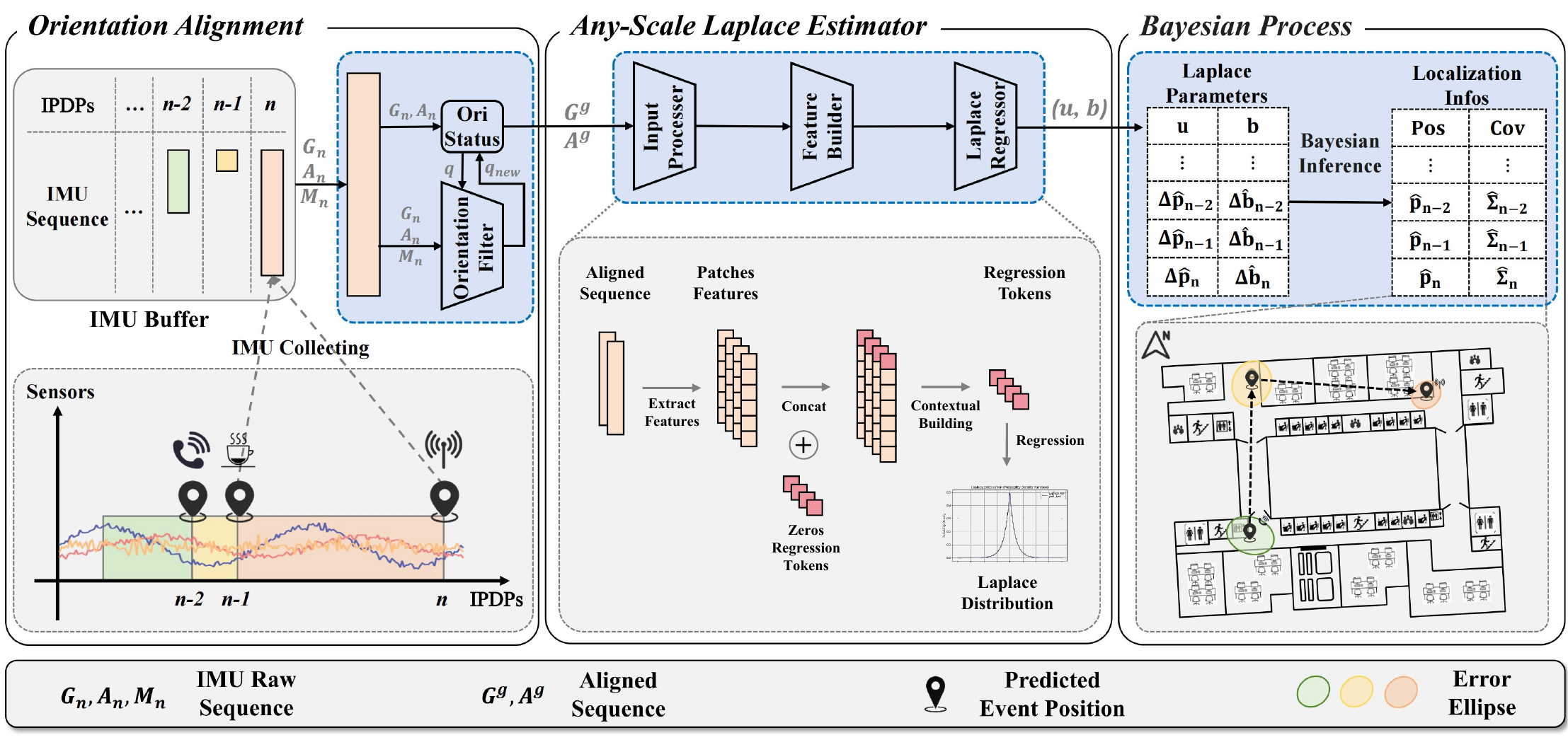}
  \caption{Overview of the ReNiL framework. The figure illustrates the ReNiL workflow in three stages.}
  \label{fig_2}
  \Description{A three-stage workflow diagram of ReNiL showing orientation alignment, any-scale Laplace estimator, and Bayesian process.}
\end{figure*}

\subsection{System Overview of ReNiL}
The workflow of ReNiL is shown in Fig.~\ref{fig_2}.
The proposed ReNiL system consists of three main stages: orientation alignment, any-scale Laplace estimator, and Bayesian process. 
(1) The orientation alignment stage continuously collects accelerations, angular rates, and magnetic fields in the background, dynamically aligning the smartphone coordinate system with the navigation coordinate system (e.g., ENU). It updates the orientation of the device through a motion-aware orientation filter. 
(2) The any-scale Laplace estimator handles aligned IMU sequences of any scale by extracting local patch features and further building contextual features. These features are combined via scale pooling into unified global features, which are subsequently used to estimate both the displacement and the uncertainty parameter modeled by a Laplace distribution. 
(3) The Bayesian process stage incorporates these displacement and uncertainty estimations into a Bayesian inference framework, sequentially refining the positioning result at each IPDP.

\subsection{Formulation for Bayesian IPDP Localization Task}

\textit{Step I:} This step is the orientation alignment stage. It ensures the directional consistency of the network's relative displacement output. 
A pedestrian inertial localization task generally involves two right-handed Cartesian coordinate systems: the navigation coordinate system and the device coordinate system. 
To avoid numerical issues and singularities, we use a quaternion, denoted by $q = w + xi + yj + zk$, to describe the rotation from the device coordinate system to the navigation coordinate system. 
For a timestamp measurement $(a,b,c)$ in the device coordinate system, and its pure quaternion $X = ai + bj + ck$, quaternion multiplication $\otimes$ is used to rotate the vector to the navigation coordinate system, as expressed by
\begin{equation}
  \begin{pmatrix} A^g \ G^g \ M^g \end{pmatrix} 
  = q^{-1} \otimes \begin{pmatrix} A \ G \ M \end{pmatrix} \otimes q.
\end{equation}
Then the quaternion $q$ is updated to a new quaternion $q_{\text{new}}$ by the orientation filter $f_o(\cdot;\cdot)$:
\begin{equation}
  q_{\text{new}} = f_o \!\left(\begin{pmatrix} A \ G \ M \end{pmatrix}, q\right).
\end{equation}

\textit{Step II:} This step constitutes the core of the Bayesian IPDPs pedestrian neural network task. 
We consider a sequence of aligned inertial measurements $x_i^g = [A_i^g, G_i^g]$ from the IPDP $t_{n-1}$ to $t_{n}$:
\begin{equation}
  X^g_{t_{n}} = \left\{x^g_{t_{n-1}}, \ldots, x^g_{t_{n}}\right\}, 
  \quad x^g_t \in \mathcal{R}^6.
\end{equation}
The goal of the localization task at IPDPs is to estimate the relative displacement at specific demand points instead of continuously estimating the trajectory. 
To quantify uncertainty at each demand point, the ground-truth displacement $\Delta p_{t_n} \in \mathcal{R}^2$ can be modeled probabilistically via a Laplace distribution:
\begin{equation}
  \Delta p_{t_n} \sim \operatorname{Laplace}\!\left(\Delta \hat{p}_{t_n}, \hat{b}_{t_n}\right),
  \quad \Delta \hat{p}_{t_n}, \hat{b}_{t_n} \in \mathcal{R}^2,
\end{equation}
where $\Delta \hat{p}_{t_n}$ is the predicted displacement, and $\hat{b}_{t_n}$ is the estimated uncertainty scale parameter at the IPDPs.

Given the aligned IMU sequence from IPDP $t_{n-1}$ to $t_{n}$, we aim to estimate the relative displacement distribution:
\begin{equation}
  (\Delta \hat{p}_{t_n}, \hat{b}_{t_n}) = f\!\left(X^g_{t_{n}}; \theta\right),
\end{equation}
where $f(\cdot; \theta)$ is a deep neural network parameterized by $\theta$, trained to output displacement and uncertainty directly, eliminating explicit velocity integration over fixed windows.

\textit{Step III:} This step describes the localization task of multiple consecutive demand points as a Bayesian localization inference framework. 
Assume that there is a series of IPDPs, expressed as
\begin{equation}
  T_d = \left\{t_0, t_1, \ldots, t_n\right\}, \quad 
  0 \leq t_0 < t_1 < \cdots < t_n \leq T.
\end{equation}
The inertial relative position estimation task corresponding to each IPDP can be described as a Bayesian inference problem. 
In the position estimation of the IPDP $t_n$, the previous step position estimation result $t_{n-1}$ is used as the prior information and jointly inferred with the network output of the current $t_n$ IPDP to obtain the posterior probability distribution of the current position:
\begin{equation}
  \mathbf{p}\!\left(p_{t_n}\right)
  = \int \mathbf{p}\!\left(p_{t_n} \mid p_{t_{n-1}}, 
  \Delta \hat{p}_{t_n}, \hat{b}_{t_n}\right)
  \mathbf{p}\!\left(p_{t_{n-1}}\right)
  d p_{t_{n-1}}.
\end{equation}
All IPDPs are connected in series to form a complete Bayesian inference chain. 
The estimation of each IPDP can clearly give the location estimate and its corresponding uncertainty, thereby achieving a unified probabilistic inference for all IPDPs.

\section{Relative Neural Inertial Locator}
\label{sec:locator}

\subsection{Orientation Alignment}
\begin{figure}[!t]
\centering
\includegraphics[width=0.47\textwidth]{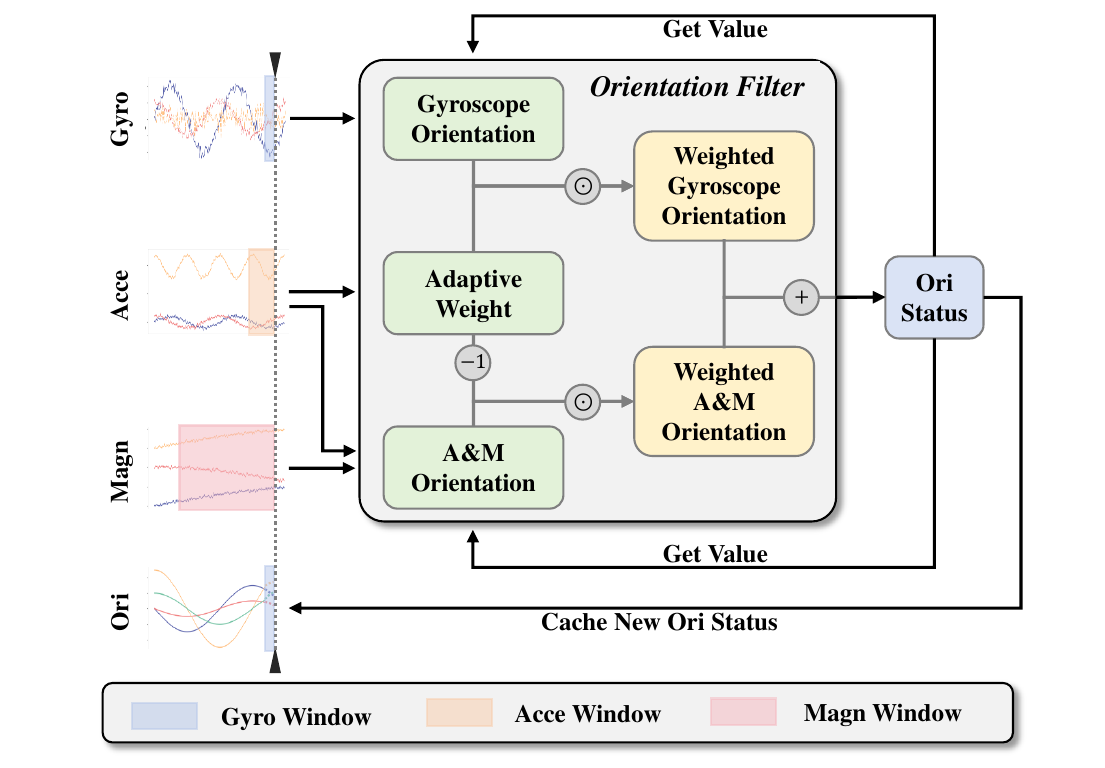}
\caption{Motion-aware orientation filter based on adaptively weighted complementary. Gyroscope, accelerometer and magnetometer are weighted and updated orientation estimation with different windows.}
\label{fig_3}
\Description{Diagram of the motion-aware orientation filter: gyroscope as the main branch, accelerometer and magnetometer each using sliding windows to compute adaptive weights; the three branches are fused via a complementary filter to update the device orientation.}
\end{figure}
Estimating the device orientation in the navigation coordinate system $\mathcal{C}$ from sequences with unknown initial state, deviation, complex motion, and interference is a challenge. Considering the periodic nature of human motion and the characteristics of three IMU sensors, we propose a window-divided update mechanism to converge the device orientation to the navigation coordinate system $\mathcal{C}$. The gyroscope serves as the primary sensor for updating the device's orientation quaternion $q$. The accelerometer and magnetometer use their own independent sliding windows to compute adaptive weights and A\&M orientation $q_a,q_m$. These adaptive weights and $q_a,q_m$ are then fused with the main gyroscope branch using a complementary filter at encounter timestamps, as shown in Fig.~\ref{fig_3}.

Calculate the adaptive weight $W_a$ for the accelerometer according to its motion intensity and orientation stability:
\begin{equation}
W_a=2\sigma(\frac{u}{|D^a|}\sum_{_{a_i \in D^a}}\|a_i-G||_2 + v\sum_{j\in \{x,y,z\}} \mathop{Var \ }\limits_{_{a_i \in D^a}}a_{ij})-1
\end{equation}
And calculate $W_m$ for the magnetometer based on the offset of the average magnetic field strength in the current magnetometer window relative to the total average magnetic field strength over the history:
\begin{equation}
W_m=2\sigma(\frac{h}{||m_p^n-\frac{1}{|D^m|}\sum_{_{m_i \in D^m}}m_{i}^n||_2})-1
\end{equation}
where $u,v,h$ is adjustable sensitivity parameters, $D^a, D^m$ is the accelerometer and magnetometer window, $\sigma(\cdot)$ is Sigmoid fuction, $m_p^n,m_{i}^n$ are the rotated average magnetic field values from history and the average magnetic field in the current magnetometer window in the navigation coordinate system, and $G$ is the gravity acceleration constant.

These adaptive weights and $q_a,q_m$ are then fused with the main gyroscope branch using a complementary filter at encounter timestamps:
\begin{equation}
q_{new}=\left\{
\begin{array}{cc}
W_a q+(1-W_a) q_a \\
\\
W_m q+(1-W_m) q_m \\
\end{array}
\right.
\end{equation}

To enhance the effectiveness of orientation correction, the window selection strategy is based on the sensor characteristics of pedestrian motion. 

For the accelerometer, the sliding window size $|D^a|$ is fixed to correspond to one walking cycle $T_{step}$, leveraging the periodic nature of human gait. 
\begin{equation}
|D^a| = T_{step}
\end{equation}
This allows the accelerometer’s orientation estimation to capture and evaluate the relatively stable motion.

The magnetometer window is dynamically adjusted according to the inferred displacement of the pedestrian. Specifically, a new window 
$D^m_{{t+1}}$ is triggered only when the estimated trajectory indicates that the user has moved beyond a specified spatial threshold $\Delta$. 
\begin{equation}
\text{If } ||p_{t+|D^m_t|} - p_{t}||_2 \geq \Delta, \text{ then open } D^m_{t+1}
\end{equation}
This distance-based window ensures that the magnetometer orientation estimation is less susceptible to transient local magnetic disturbances, as the spatial decorrelation allows for a more stable extraction of the Earth’s magnetic field direction.

\subsection{ASLE Structure}

\begin{figure*}[!t]
\centering
\includegraphics[width=0.95\textwidth]{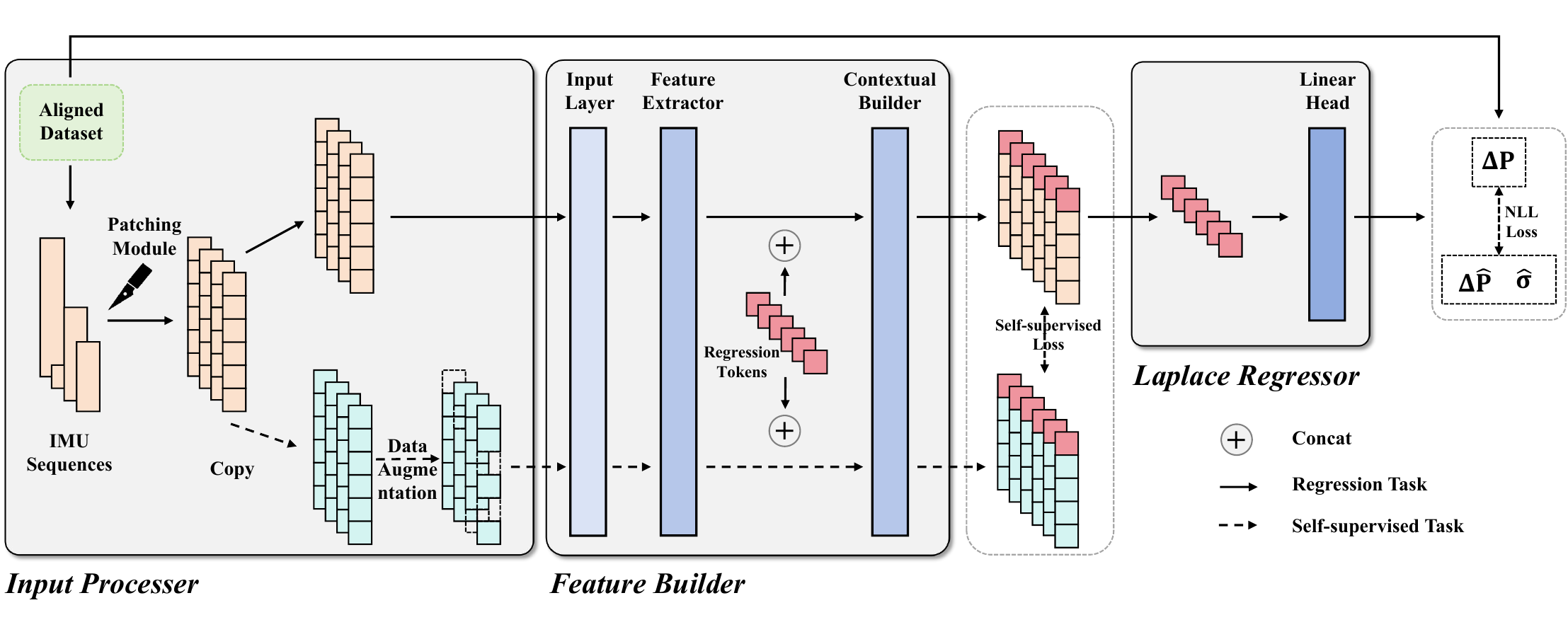}
\caption{The structure and training tasks of the ASLE. The solid line represents the Bayesian regression task, and the loss is calculated by the label and the network output. The dashed line represents the self-supervised task, and the loss is calculated by the output of the contextual builder module for the input with data augmentation and without it.}
\label{fig_4}
\Description{Block diagram of ASLE showing five modules: patching, input, feature extractor with 1D residual blocks, contextual builder with multi-head self-attention and a regression token, and an MLP regressor that outputs Laplace parameters for displacement and uncertainty; arrows distinguish supervised regression and self-supervised feature-matching paths.}
\end{figure*}
We propose an any-scale Laplace estimator (ASLE), which consists of five key components: 1) the patching module, 2) the input module, 3) the feature extractor, 4) the contextual builder, and 5) the regression module. Each module within the ASLE can be replaced with other general modules that perform the same function. In this paper, we have designed a specific verification model for the system. The structure of the ASLE is shown in Fig.~\ref{fig_4}.

\paragraph{Patching module}  
The aligned acceleration $A^g \in \mathcal{R}^{B \times T \times 3}$ and angular rates $G^g \in \mathcal{R}^{B \times T \times 3} $ data are concatenated to a 2D tensor $X \in \mathcal{R}^{B \times T \times 6}$, where $T$ represents the total number of data samples collected during two IPDPs and $B$ is the mini-batch number.
Next, the patch operation deviates $X$ into a 3D tensor $X \in \mathcal{R}^{B \times P \times 6 \times L}$, where $P$ is the number of patches, $L$ is the patch length, and the tail patch that does not reach $L$ is padded with zeros. The number of patches is calculated as:
\begin{equation}
P = \lceil \frac{T}{L} \rceil
\end{equation}
A larger patch length $L$ increases network efficiency, but it requires a stronger feature extraction capability since each patch will contain richer information.

\paragraph{Input module}  
The input module is used to map the original 6 features of the input \( X \in \mathcal{R}^{B \times P \times 6 \times L} \) to a higher-dimensional channels and compress the patch length to greatly reduce computation overhead. Specifically, a typical convolutional input layer is applied to embed \( X \) from 6 channels into a higher channels and L into a shorter length:
\begin{equation}
X^e \in \mathcal{R}^{B \times P \times e \times L'}
\end{equation}
where \( e \) is the channels dimension and $L'$ is the compressed length.

\paragraph{Feature extractor} The 1D version of ResNet18~\cite{he2016deep} has demonstrated strong sequence feature extraction capabilities in previous work. We use its improved residual block (replace BN with GN) as a feature extractor to capture motion pattern features in each patch in parallel. Specifically, for $X^e$, multiple 1D residual blocks are applied sequentially to extract patch features. Each residual block consists of two convolutional layers and two GN layers. The number of channels in each residual block is double that of the previous block. The module contains a total of \( N_1 \) residual blocks. For each block, the convolutional layer has a stride of 2 for first layer, 1 for others and a kernel size of 3. The output of each 1D residual block serves as the input for the next 1D residual block. The outputs of the final residual block are concatenated along the channel axis. The output of this module is: 
\begin{equation}
X \in \mathcal{R}^{B \times P \times C \times W} 
\end{equation}
where \( C \) is the number of output channels from the last residual block, and \( W \) represents the extracted one-dimensional feature.

\paragraph{Contextual builder} The multi-head self-attention is highly effective for context modeling and able to handle sequences of any scale. It also offers fast computation, making it an excellent choice for modeling movement context between patches. Specifically, to allow the attention layer to model features in the last two dimensions, we flatten the second and third axes of the output tensor \( X \) from the 1D feature extractor, and add an additional zeros regression tokens to the time dimension. This yields a new tensor 
\begin{equation}
    X \in \mathcal{R}^{B  \times (P+1) \times (C\times W)}
\end{equation}
Then, \( N_2 \) multi-head self-attention blocks are applied to build the temporal local context of the features, resulting in a robust features map. Each multi-head self-attention block works as follows: 
\begin{equation}
\mathrm{MHA}(X) = 
\Big(
\mathrm{softmax}\!\left(
\frac{(X W_Q)(X W_K)^{\top}}{\sqrt{d}}
\right)
(X W_V) + X
\Big) W_O
\end{equation}
Where $W_Q,W_K,W_V, W_O$ are learnable projection matrices, d is attention scaling dimension.
Finally, we obtain robust features map:
\begin{equation}
    X \in \mathcal{R}^{B  \times (P+1) \times D}
\end{equation}
Where $D$ is the attention dimension. This operation aims to fuse the contextual potential relationships between patches, extract the displacement information of the entire IMU into regression tokens, and improve the robustness of the final feature map.

\paragraph{Regression module} The regression module aims to regress the inherent features in regression tokens into Laplace parameter outputs. Specifically, a multilayer perceptron (MLP) $O(\cdot)$, which contains fully connected, dropout, and ReLU layers, is applied to regress $\Delta \hat{p}, \hat{b}$.
\begin{equation}
    (\Delta \hat{p}, \hat{b}) = O(X),\Delta \hat{p}, \hat{b}\in R^{B \times 2}
\end{equation}
where $\Delta \hat{p}, \hat{b}$ are the predicted relative displacement and Laplace scale parameters.

\subsection{Joint Training with Uncertainty Estimation}

In general, our optimization objective $\mathcal{L}$ is defined as:
\begin{equation}
\mathcal{L} =\mathcal{L}_{nll} + \mathcal{L}_{fm}    
\end{equation}
where $\mathcal{L}_{NLL}$ and $\mathcal{L}_{fm}$ correspond to the supervised and unsupervised components of the framework respectively.

\paragraph{Supervision} In the two-dimensional displacement regression problem, we estimate the parameters of the neural network $\theta$ using maximum likelihood, which involves maximizing the likelihood estimate:
\begin{equation}
	\mathop{\arg\max}_{\theta} \prod \limits_{d_i \in \mathcal D}
    \mathbf{p}_\theta(\Delta{p}_i|d_i)
\end{equation}
where $\mathcal D$ is the sample set, $\Delta{p}_i$ is the ground truth displacement of sample $d_i$. 

Specifically, we assume that \( \Delta{p}_i \) follows a Laplace distribution: \( \Delta {p}_i \sim \operatorname{Laplace}(\Delta\hat{p}_i, \hat{b}_i) \). To facilitate backpropagation, we transform the maximum likelihood estimation problem into minimizing the negative log-likelihood (NLL):
\begin{equation}
\mathcal{L}_{nll}=-\log\prod \limits_{d_i \in \mathcal D} \left(\frac{1}{2\hat{b}_i} \exp \left(-\frac{|\Delta p_i-\Delta\hat{p}_i|}{\hat{b}_i}\right)\right)
\label{eq22}
\end{equation}

We also need to consider another training optimization problem. Displacements on different scales are physical quantities with large fluctuations. Directly regressing the displacements is not only prone to gradient vanishing or exploding problems, but also requires a more complex structure and design for the output layer.
Now consider the average speed \( {v}_i \) as a random variable, has the following properties:
\begin{equation}
E(\Delta p_i) = t_i E( {v}_i),  D(\Delta p_i) = t_i^2 D( {v}_i)
\label{eq23}
\end{equation}
Probability and statistics theory provides a rigorous transformation relationship between ${v}_i, \Delta p_i$. So, the neural network regresses the average speed distribution $\hat{v}_i,\hat{b}_{vi}$ rather than the displacement and can easily reproduce $\Delta \hat{p}_i,\hat{b}_i$ by Eq.\ref{eq23}. 

By combining Eq.\ref{eq22} and Eq.\ref{eq23}, we simplify the \( \operatorname{L_{NLL}} \) function to make it more numerically stable. The resulting form is:
\begin{equation}
\mathcal{L}_{nll}=\sum_{d_i \in \mathcal D}(\frac{|v_i-\hat{v}_i|}{e^{\log \hat{b}_{vi}}}+\log \hat{b}_{vi}+\log t_i)+Cst
\end{equation}

\paragraph{Self-supervision} To enhance the context modeling mechanism of ASLE and obtain more robust spatio-temporal motion features, we propose a self-supervision method aimed at improving the model's generalization ability and its resistance to false information. 

Specifically, for the original input $X\in R^{B\times P\times 6\times L}$ of the network, we record its features map $X_f \in \mathcal{R}^{B  \times (P+1) \times D}$, which is the output after the contextual builder. We then perform a series of data augmentations on the original input $X$, including:
\begin{itemize}
\item Partial masking: Randomly select patches from the input tensor and mask all values with 0.
\item Quaternion constant bias interference: Apply a small random 3D rotation disturbance to the entire input tensor without altering the label values.
\item Gaussian noise: Add Gaussian noise with a mean of 0 to the acceleration and gyroscope data to simulate sensor bias.
\item Heading rotation: Apply a 2D rotation on the same navigation plane to the entire input tensor and its label values.
\item Abnormal protrusions: Add several smooth protrusions as disturbances along each signal axis of the input tensor.
\end{itemize}

Subsequently, we obtain the features map $\hat{X}_f \in \mathcal{R}^{B  \times (P+1) \times D}$ of the contextual builder output. Finally, we calculate the MSE loss $L_{fm}$ between $\hat{X}_f$ and $X_f$ to construct robust context modeling features:
\begin{equation}
\mathcal{L}_{fm}=\sum_{d_i \in \mathcal D}(\hat{X}_{fi} -X_{fi} )^2
\end{equation}

\subsection{Bayesian Process}
\begin{figure*}[!t]
\centering
\includegraphics[width=0.95\textwidth]{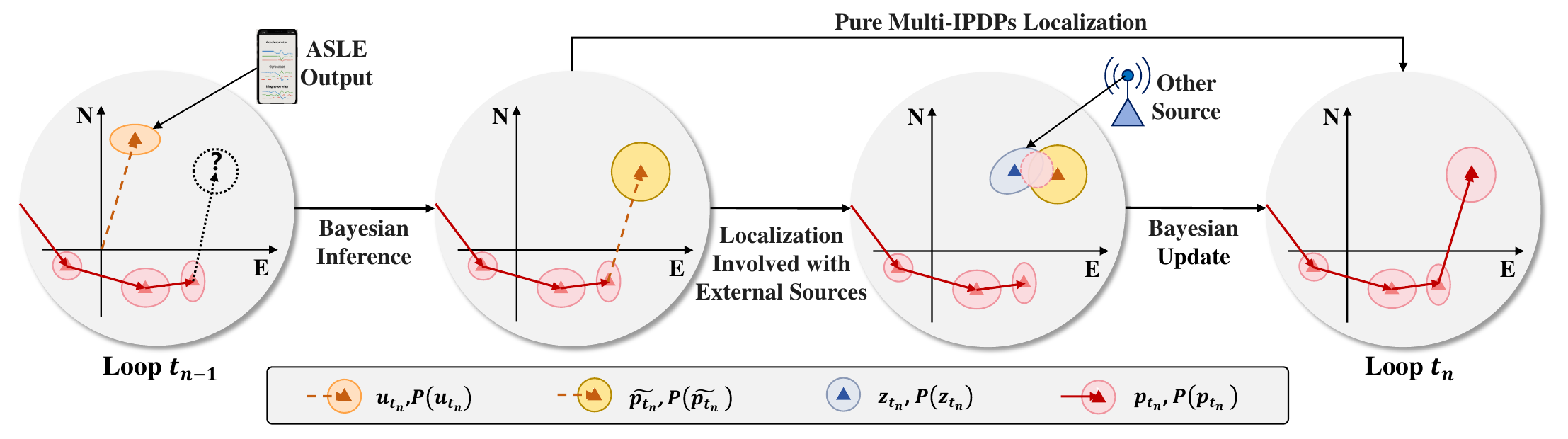}
\caption{Bayesian process for pure multi-IPDPs localization or involved with external sources. ReNiL uses Bayesian inference to combine ASLE outputs of IPDP $t_n$ with prior position \(p_{t_{n-1}}\) of IPDP $t_{n-1}$. For pure multi-IPDPs localization, ReNiL produces final posterior estimates \(p_{t_n}\) directly;  For multi-IPDPs localization involved with external sources, ReNiL updates Bayesian inference result with the external observations \(z_{t_{n}}\) to produce final posterior estimates \(p_{t_{n}}\). Then move to the next loop.}
\label{fig_16}
\Description{Flow diagram of the Bayesian process: prior at IPDP t_{n-1} combined with ASLE outputs to produce posterior at t_n; optional external observations are fused via a Bayesian filter; the chain iterates across IPDPs.}
\end{figure*}
The output of ASLE is the relative displacement and scale parameters between two IPDPs. However, our localization inference is a continuous process. Next, we will use Bayesian inference to connect all IPDPs in series to form a complete Bayesian inference chain. In the case of IPDPs involved with other external localization sources, we use a Bayesian filter~\cite{chen2003bayesian} for loose coupling.

First, we need to define the initial status and its distribution as $p_0$ and $\mathbf{p}(p_0)$, respectively.
And define the state transition model, which describes the change from time $t_{n-1}$ to time $t_n$:
\begin{equation}
    p_{t_n}=f\left(p_{t_{n-1}}, u_{ t_n}\right)+w_{ t_n}
    \label{eq27}
\end{equation}
where $f(\cdot)$ is the state transition function, $u_{ t_n}$ is the ASLE output (control vector) and $w_{ t_n}$ represents ASLE process noise.

\paragraph{For pure multi-IPDPs localization} 
We regard the status ${p}_{t_{n-1}}, \mathbf{p}\left({p}_{t_{n-1}} \mid u_{t_0:t_{n-1}}\right)$ of IPDP $t_{n-1}$ as a priori information, and combine it with the output of ASLE. According to Bayes' theorem, we can infer the posterior probability of IPDP $t_{n}$ status:
\begin{equation}
\mathbf{p}\left({p}_{t_n} \mid u_{t_0:t_{n}}\right)= \int \mathbf{p}\left({p}_{t_n} \mid {p}_{t_{n-1}},u_{t_{n}} \right) \mathbf{p}\left({p}_{t_{n-1}} \mid u_{t_0:t_{n-1}}\right) d {p}_{t_{n-1}}
\label{eq28}
\end{equation}
Combining the inference Eq.\ref{eq28} and the status transition Eq.\ref{eq27}, we can get the complete Bayesian IPDPs status inference chain, as shown in Fig.~\ref{fig_16}.

\paragraph{For multi-IPDPs localization involved with external sources} 
We use a Bayesian filter for loose coupling, which iteratively improves the estimation accuracy of unknown states by predicting and updating the fusion of information from different observation sources. 

Define the observation model, which describes mapping from the state space to the observation space:
\begin{equation}
    z_{t_n}=h\left(p_{t_n}\right)+v_{t_n}
    \label{eq29}
\end{equation}
where $h(\cdot)$ is the observation function and $v_{t_n}$ is the observation noise.

\textit{Step 1}: In the inference phase, the system motion model is used to make forward predictions of the state distribution:
\begin{equation}
\mathbf{p}\left(\widetilde{p_{t_n}} \mid {z}_{t_0: t_{n-1}}, u_{t_0:t_{n}}\right) = 
\int \mathbf{p}\left(p_{t_n} \mid p_{t_{n-1}}, {u}_{t_n}\right)\, \mathbf{p}\left(p_{t_{n-1}} \mid {z}_{t_0: t_{n-1}}, {u}_{t_{0}:t_{n-1}}\right) \, d p_{t_{n-1}}
\label{eq30}
\end{equation}

\textit{Step 2}: In the update phase, the predicted state distribution is corrected using the sensor observation at the current IPDP $t_{n}$:
\begin{equation}
\mathbf{p}\left(p_{t_n} \mid {z}_{t_0: t_{n}}, u_{t_0:t_{n}}\right) \propto \mathbf{p}\left({z}_{t_n} \mid p_{t_n}\right) \mathbf{p}\left(\widetilde{p_{t_n}} \mid {z}_{t_0: t_{n-1}}, u_{t_0:t_{n}}\right)
\label{eq31}
\end{equation}

Combining the observation model Eq.\ref{eq29}, the inference Eq.\ref{eq30}, the update Eq.\ref{eq31} and the status transition Eq.\ref{eq27}, we can get the complete Bayesian IPDPs status inference chain involved with external sources, as shown in Fig.~\ref{fig_16}.

\section{Experimental Evaluations}
\label{sec:evaluation}

\subsection{Experiment Setup}
We have performed a comprehensive set of experiments to thoroughly evaluate the entire system. This includes comparing the system's performance with baseline methods on a publicly available dataset RoNIN-ds and our own WUDataset.

\paragraph{Baseline}
For displacement-regression experiments we employ ASLE-ns, a family of our ASLE that accepts fixed n-second input windows. Because the competing baselines PDR, RoNIN-ResNet18 (RoRes18), TLIO, CTIN, and iMoT can each operate only at a single preset window size, we instantiate four representative variants (n = 1, 5, 10, 20) to span short, medium, and longer motion segments; these values were chosen simply to give broad temporal coverage rather than for any dataset-specific reason. We evaluate all methods head-to-head and then carry forward the two strongest baselines for deeper analysis. For device orientation estimation, our motion-aware orientation filter is benchmarked against the classical Mahony~\cite{mahony2008nonlinear} and Madgwick~\cite{madgwick2011estimation} algorithms.

\paragraph{Dataset}
In pedestrian inertial navigation research using neural networks, the open-source RoNIN-ds dataset is widely recognized. To further evaluate the applicability of ReNiL, we collected the WUDataset, which contains various pedestrian motion data collected from both complex indoor and outdoor environments.

The WUDataset collection was carried out using Google Pixel 6A, Huawei mate 40E, Iphone 16 Pro and Xiaomi 15 smartphone (IMU sampling rate: 200Hz), a Vicon optical system (sampling rate: 100Hz), and a LiDAR-SLAM system (sampling rate: 200Hz), all synchronized through a custom-developed Android application. The WUDataset is organized as follows:

\begin{itemize}
\item Indoor: An 8m × 20m space observed by the Vicon optical system.
\item Outdoor: Various outdoor environments such as college halls, corridors, and open areas rebuilt by a LiDAR-SLAM system.
\item Subjects: The WUDataset includes 120 independent trajectories recorded by 28 participants (17 men and 11 women, with heights ranging from 1.5 to 1.9 meters), covering a total distance of more than 40 kilometers. 
\item Movement patterns: categorized into three intensity levels of exercise: slow walking, normal walking speed, and running. In addition, multiple equipment positions were used during data collection, including hand hold, shoulders, arms, front and back pockets, backpacks, handbags, swinging arms, and chest pockets.
\item Dataset division: Both the WUDataset and RoNIN-ds datasets are divided into training, validating, and testing sets in a 5:1:4 ratio. The test set is further subdivided into seen and unseen subject groups. Note that the division of RoNIN-ds is consistent with the RoNIN open-source method.
\end{itemize}

\paragraph{Evaluation metrics} TABLE \ref{tab:metrics}
\begin{itemize}
\item MAE: Mean absolute error between displacement predictions and ground truth.
\item ADE: Average displacement error per second.
\item HE: Heading error between position predictions and ground truth.
\item QAE: Quaternion angle error, quantifying orientation difference in quaternion form.
\item CS: Cosine similarity between two quaternions.
\end{itemize}
\begin{table}[!t]
\caption{Evaluation Metrics and Definitions}
\label{tab:metrics}
\centering
\footnotesize
\setlength{\tabcolsep}{8pt}
\renewcommand{\arraystretch}{1.2}
\begin{tabular}{@{}ll@{}}
\toprule
\textbf{Metric} & \textbf{Formula} \\
\midrule
MAE
  & $\displaystyle \mathrm{MAE}=\frac{1}{N}\sum_{i=1}^{N}\big\|\hat{\boldsymbol{p}}_{i}-\boldsymbol{p}_{i}\big\|_{1}$ \\[2pt]
ADE 
  & $\displaystyle \mathrm{ADE}=\frac{1}{N}\sum_{i=1}^{N}\frac{\big\|\Delta\hat{\boldsymbol{p}}_{i}-\Delta\boldsymbol{p}_{i}\big\|_{2}}{\Delta t_{i}}$ \\[2pt]
HE
  & $\displaystyle \mathrm{HE}=\frac{1}{N}\sum_{i=1}^{N}\big|\hat{\theta}_{i}-\theta_{i}\big|$ \\[2pt]
QAE 
  & $\displaystyle \mathrm{QAE}=\frac{1}{N}\sum_{i=1}^{N}2\,\arccos\!\big(\,|\langle\hat{\boldsymbol{q}}_{i},\boldsymbol{q}_{i}\rangle|\,\big)$ \\[2pt]
CS
  & $\displaystyle \mathrm{CS}=\frac{1}{N}\sum_{i=1}^{N}\frac{\langle\hat{\boldsymbol{q}}_{i},\boldsymbol{q}_{i}\rangle}{\|\hat{\boldsymbol{q}}_{i}\|\,\|\boldsymbol{q}_{i}\|}$ \\
\bottomrule
\end{tabular}
\end{table}

\subsection{Implementation Details}

\paragraph{Orientation alignment} The sliding window size for the accelerometer is set to 1 second, and the $\Delta$ of the magnetometer is set to 10 meters. The parameters \( u \), \( v \), and \( h \) are set to 1, 1000, 8, and the gravity acceleration constant \( G \) is set to 9.81 m/s\(^2\). Note that all these parameters can be adjusted according to the application scenario.

\paragraph{ASLE details}
We implemented ASLE using the PyTorch 2.5.1~\cite{pytorch251} framework on a server equipped with an Intel Xeon Silver 4210R CPU and a GeForce GTX 3090 GPU. The model utilizes 200Hz IMU input, with a patch length of 200. The channel, kernel size, and stride of the input module are set to 32, 7, 2. The feature extractor consists of four 1D residual layers with output channels of 32, 64, 128, and 256, respectively. The context builder consists of 4 MHA blocks with 4 heads, and 64 dimensions for each patch. The regression module is conducted as two fully connected layers with hidden size 128, ReLU layer, and the output is a vector of length 4. Adam~\cite{kingma2014adam} is used as the optimizer, with the learning rate scheduler set to ReduceLROnPlateau~\cite{pytorch_reduce_lr_on_plateau}, using a factor of 0.1 and patience of 10. The initial learning rate for the network parameters is set to \( 1 \times 10^{-4} \), with a minimum learning rate of \( 1 \times 10^{-8} \). The mini-batch size is 32, and the model is trained for a total of 200 epochs.

\paragraph{Bayesian process}

We use the Bayesian inference strategy of Kalman filtering combined with Gibbs sampling to form a linear dynamic system that handles Laplace process noise. We use the process and observation models of Eq.\ref{eq27}, Eq.\ref{eq29} and implement Eq.\ref{eq28}, Eq.\ref{eq30}, and Eq.\ref{eq31}.

\textit{For pure multi-IPDPs localization.} We consider it as a Gaussian-exponential mixture. For each step $t_n$, introduce the hidden variable $\tau_{t_n}$:
\begin{equation}
\tau_{t_n} \sim \operatorname{Exp}(\frac{1}{2\boldsymbol{\Sigma}_b^2})
\end{equation}
where $\boldsymbol{\Sigma}_b=\operatorname{diag}\left(\Delta t_n b_{v_x},\Delta t_n b_{v_y}\right)$.
So the Laplace distribution noise can be conditionally Gaussian representation into a Gaussian distribution noise:
\begin{equation}
{w}_{t_n} \mid \tau_{t_n} \sim \mathcal{N}\left({0}, \tau_{t_n} \boldsymbol{\Sigma}_w\right)
\label{eq33}
\end{equation}
where $\boldsymbol{\Sigma}_w=\operatorname{diag}\left(2\left(\Delta t_n b_{v_x}\right)^2,2\left(\Delta t_n b_{v_y}\right)^2\right)$. 
Then, we use $\tau_{t_n} \boldsymbol{\Sigma}_w$ to construct a covariance matrix that conforms to the Gaussian distribution to complete the Bayesian inference chain.

\textit{For multi-IPDPs localization involved with external sources.} We also introduce $\tau_{t_n}$, but it is updated using Gibbs according to the Rao-Blackwell theorem. Calculate the sufficient statistic Mahalanobis distance $\delta_{t_n}$, and the posterior probability of $\tau_{t_n}$:
\begin{equation}
\delta_{t_n}=\left(p_{t_{n-1}}-p_{t_n}\right)^{\top} \boldsymbol{\Sigma}_w^{-1}\left(p_{t_{n-1}}-p_{t_n}\right)
\end{equation}
\begin{equation}
\mathbf{p}\left(\tau_{t_{n}} \mid \delta_{t_n}\right)=\operatorname{InvGauss}\left(\mu=\sqrt{\delta_{t_n}}, \lambda=\delta_{t_n}\right)
\end{equation}
Then, resampling $\tau_{t_n}$ based on $\mathbf{p}\left(\tau_{t_{n}} \mid \delta_{t_n}\right)$. We conditionally Gaussian represent the Laplace distribution noise in a Gaussian distribution noise as in Eq.\ref{eq33}. The rest of the inference and update phases are just like a standard Kalman filter.

\subsection{Experimental Results of ReNiL Framework}

\paragraph{\textbf{Displacement accuracy performance}}

\begin{table*}[!t]
\centering
\caption{Displacement accuracy on RoNIN-ds and WUDataset. Top-1 \best{bold red}, Top-2 \second{bold blue}, Top-3 \third{bold green}. Ties share the same rank.}
\label{table_3}
\small
\setlength{\tabcolsep}{2pt}
\renewcommand{\arraystretch}{1.2}
\begin{tabular}{lllrrrrr|rrrr}
\toprule
\textbf{Dataset} & \textbf{Subjects} & \textbf{Metric $\downarrow$}  & \textbf{PDR} & \textbf{RoRes18} & \textbf{TLIO} & \textbf{CTIN} & \textbf{iMoT} & \textbf{ASLE-1s} & \textbf{ASLE-5s} & \textbf{ASLE-10s} & \textbf{ASLE-20s} \\
\midrule
\multirow{6}{*}{RoNIN-ds}
 & \multirow{3}{*}{Seen} & MAE &
  28.24 & \third{4.93} & \second{4.84} & 5.19 & 5.24 & 5.12 & 4.67 & \best{4.62} & 4.67 \\
 &  & ADE &
  0.36 & \second{0.11} & \second{0.11} & 0.13 & 0.13 & 0.13 & 0.06 & 0.05 & \best{0.05} \\
 &  & HE &
  1.27 & 0.77 & \second{0.76} & \second{0.76} & 0.79 & 0.75 & 0.61 & 0.56 & \best{0.41} \\
 & \multirow{3}{*}{Unseen} & MAE &
  32.81 & \third{6.97} & \second{6.86} & 7.23 & 7.19 & 6.95 & 6.26 & 6.30 & \best{6.24} \\
 &  & ADE &
  0.38 & \second{0.11} & \second{0.11} & 0.13 & 0.13 & 0.18 & 0.10 & 0.08 & \best{0.07} \\
 &  & HE &
  1.31 & \second{0.77} & 0.83 & \third{0.82} & \third{0.82} & 0.75 & 0.61 & 0.52 & 0.40 \\
\midrule
\multirow{6}{*}{WUD}
 & \multirow{3}{*}{Seen} & MAE &
  21.63 & \third{3.08} & \second{2.97} & 3.16 & 3.38 & 3.28 & 2.65 & 2.18 & \best{1.90} \\
 &  & ADE &
  0.51 & \second{0.22} & \third{0.23} & 0.25 & 0.26 & 0.27 & 0.11 & 0.07 & \best{0.04} \\
 &  & HE &
  1.16 & \third{0.81} & \second{0.80} & \third{0.81} & 0.82 & 0.83 & \best{0.58} & 0.60 & 0.61 \\
 & \multirow{3}{*}{Unseen} & MAE &
  26.94 & \third{2.37} & \second{2.22} & 2.41 & 2.57 & 2.38 & 2.04 & 1.72 & \best{1.72} \\
 &  & ADE &
  0.49 & \second{0.13} & \second{0.13} & 0.15 & 0.16 & 0.15 & 0.07 & 0.05 & \best{0.03} \\
 &  & HE &
  1.18 & \third{0.51} & \second{0.49} & 0.53 & 0.54 & 0.53 & 0.35 & \best{0.32} & 0.61 \\
\bottomrule
\end{tabular}
\end{table*}

The purpose of this experiment is to evaluate the displacement accuracy of the ASLE network compared to several baseline methods, tested on both seen and unseen subjects in the RoNIN-ds and WUDataset datasets. As shown in Table~\ref{table_3}, when limited to a fixed time scale of  1 second, the displacement accuracy of all methods remains relatively similar across both datasets. When the time scale restriction is removed, the distinctive advantage of ASLE becomes evident. Its inherent design to process variable-scale IMU sequences allows it to outperform other methods that rely on fixed-length sliding windows. This flexibility enables ASLE to effectively capture temporal dependencies, which translates into substantially improved displacement accuracy, particularly pronounced on the more complex WUDataset. For the RoNIN-ds dataset, where motion patterns tend to be simpler and more repetitive, the accuracy improvements offered by ASLE are less dramatic. This is likely because existing baseline models already perform sufficiently well in mapping motion features to displacement within the context of the dataset. However, ASLE still achieves measurable gains, reinforcing its generalization capability. In general, these results highlight the effectiveness of ASLE in reducing displacement errors and its robustness to variations in input sequence length and complexity. 


\paragraph{\textbf{Alleviation of integration error}}
\begin{figure*}[!t]
\centering
\includegraphics[width=0.95\textwidth]{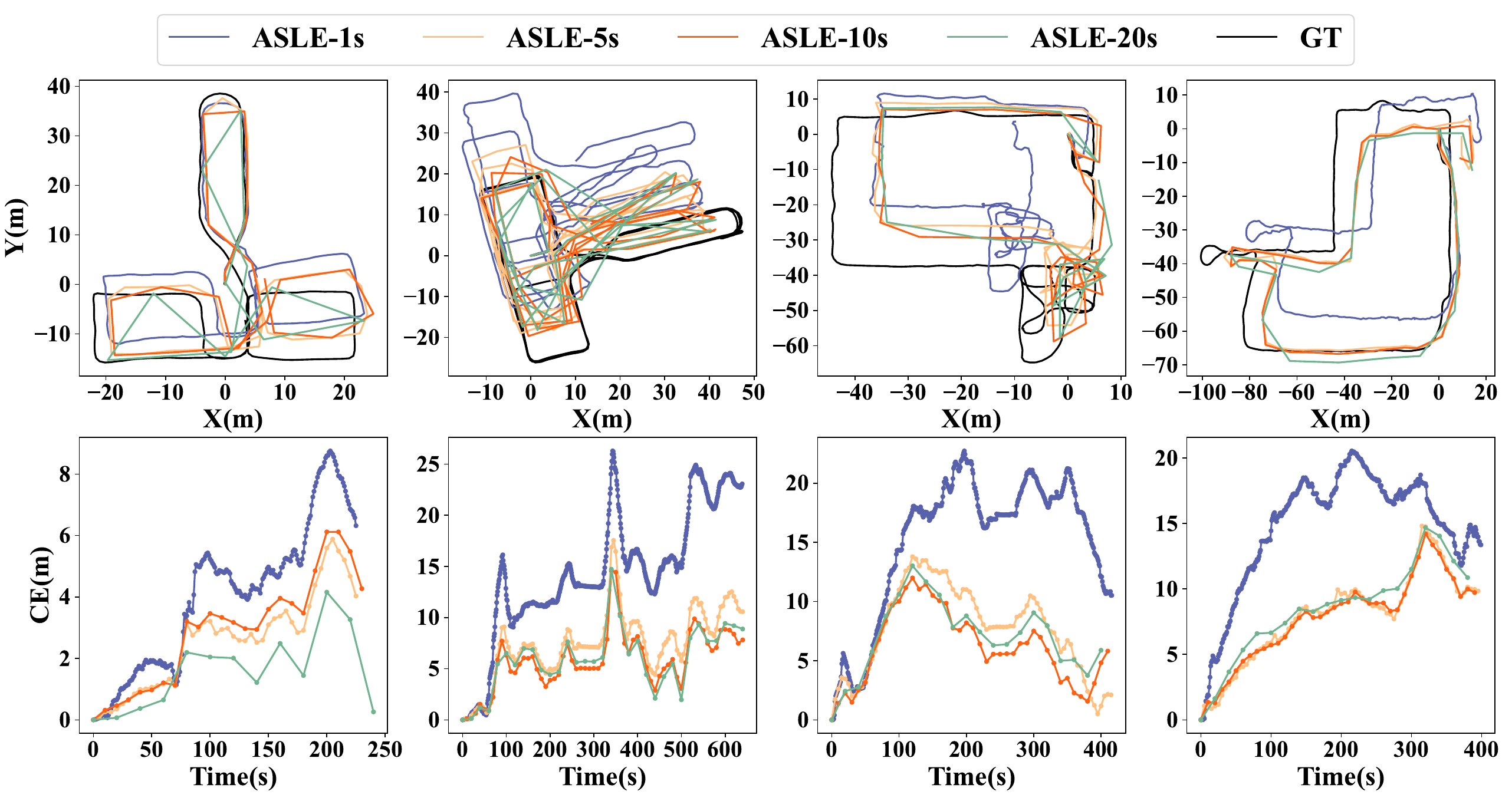}
\Description{Four representative cases from RoNIN-ds and WUDataset: top row compares multi–time-scale position estimates; bottom row shows cumulative MAE over time.}
\caption{Four typical cases from RoNIN-ds and WUDataset. The top figure of each column is a direct comparison of position estimation at different time scales, and the bottom figure shows the change of cumulative error over time.}
\label{fig_6}
\end{figure*}

This experiment verifies the ability of ASLE to reduce cumulative displacement errors arising from IMU sequence integration. Fig.~\ref{fig_6} presents four typical cases from the RoNIN-ds and WUDataset. The upper row visualizes position trajectories estimated at multiple time scales, while the lower row illustrates the corresponding MAE evolution over time. In Cases I and II, ASLE demonstrates robustness against sudden error spikes that frequently occur in fixed time-scale estimates (e.g. PLAS-1s), with longer time scales effectively smoothing these fluctuations. Cases III and IV highlight the superiority of ASLE over the fixed 1s window in compensating for cumulative errors caused by pedestrian step length variations and subtle orientation drifts. The mutual verification mechanism between patches enables more accurate and stable relative displacement estimations, especially evident in the reduced deviations from the ground truth.
\begin{figure}[!t]
\centering
\includegraphics[width=0.47\textwidth]{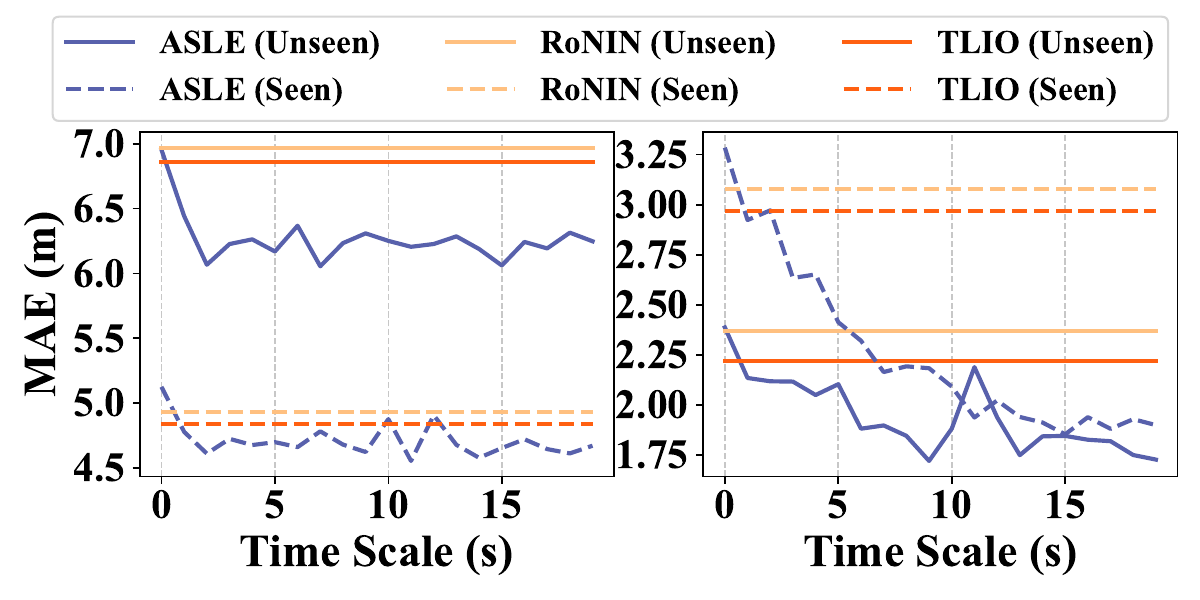}
\Description{MAE of ASLE, RoRes18, and TLIO across different time scales on RoNIN-ds (left) and WUDataset (right).}
\caption{The performance of ASLE, RoRes18, and TLIO at different time scales. The left side is the RoNIN-ds, and the right side is the WUDataset.}
\label{fig_7}
\end{figure}
Further quantitative analysis in Fig.~\ref{fig_7} compares ASLE’s MAE performance with RoRes18 and TLIO across varying time scales for both seen and unseen subjects. Although ASLE’s performance at small time scales is comparable to baseline methods, it rapidly surpasses them as the time scale extends, demonstrating marked improvements in long-term displacement accuracy. Notably, on the WUDataset with more complex human motions, ASLE achieves a reduction of 40\% in MAE for seen subjects, demonstrating its capacity to effectively capture long-range temporal dependencies. These results validate ASLE can significantly mitigates integration errors. 

\paragraph{\textbf{Anti-interference performance}}  
\begin{figure}[!t]
\centering
\includegraphics[width=0.47\textwidth]{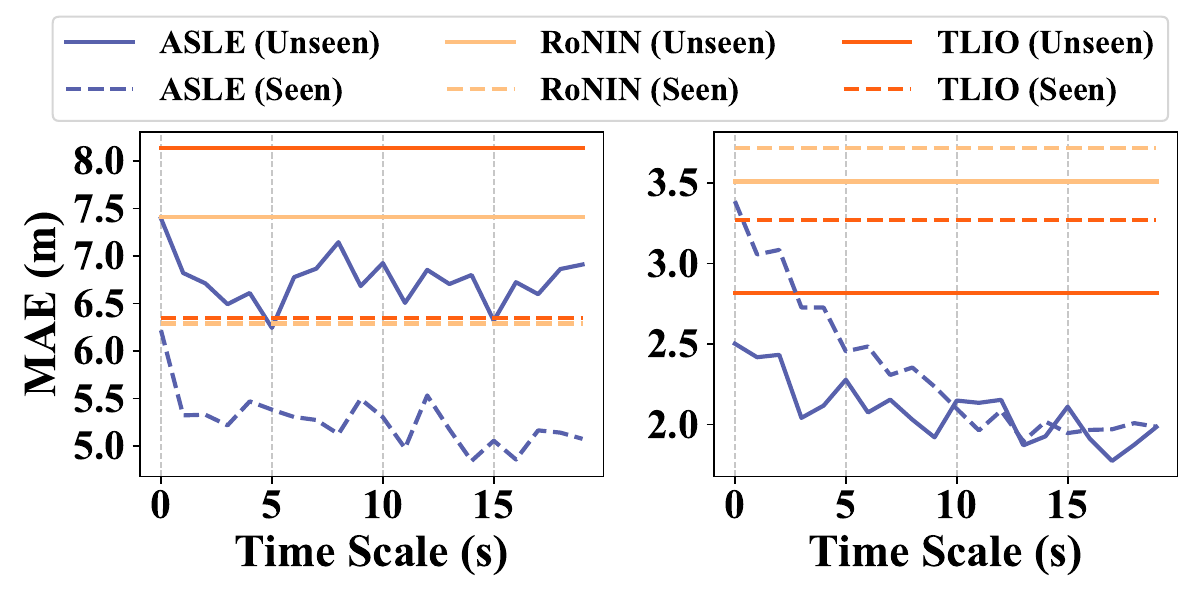}
\Description{MAE comparison under injected IMU noise, random orientation perturbations, and random data dropouts for ASLE, RoRes18, and TLIO on noisy RoNIN-ds and WUDataset.}
\caption{The performance of ASLE, RoRes18 and TLIO at different time scales. The left side is the noisy RoNIN-ds, and the right side is the noisy WUDataset.}
\label{fig_8}
\end{figure}
This experiment evaluates the robustness of ASLE against severe sensor interference commonly encountered in real-world IMU systems. The interference was simulated by injecting Gaussian noise into the 6-axis IMU signals, introducing random quaternion orientation perturbations, and randomly dropping data segments. Fig.~\ref{fig_8} presents the MAE of ASLE, RoRes18 and TLIO over varying time scales on the noisy RoNIN-ds and WUDataset. It is observed that both baselines suffer rapid and significant degradation in displacement accuracy as noise levels increase. In contrast, ASLE shows only modest error increases at short time scales and maintains stable and low MAE at longer time scales, closely matching its performance on noise-free data. These performance gains are due to ASLE’s use of context-aware sequence modeling and mutual feature verification across patches, which enable it to effectively suppress the influence of transient noise and missing data. This design ensures robustness against common IMU disturbances, resulting in a more reliable displacement estimation.

\paragraph{\textbf{Laplace distribution performance}}
\begin{figure}[!t]
\centering
\includegraphics[width=0.47\textwidth]{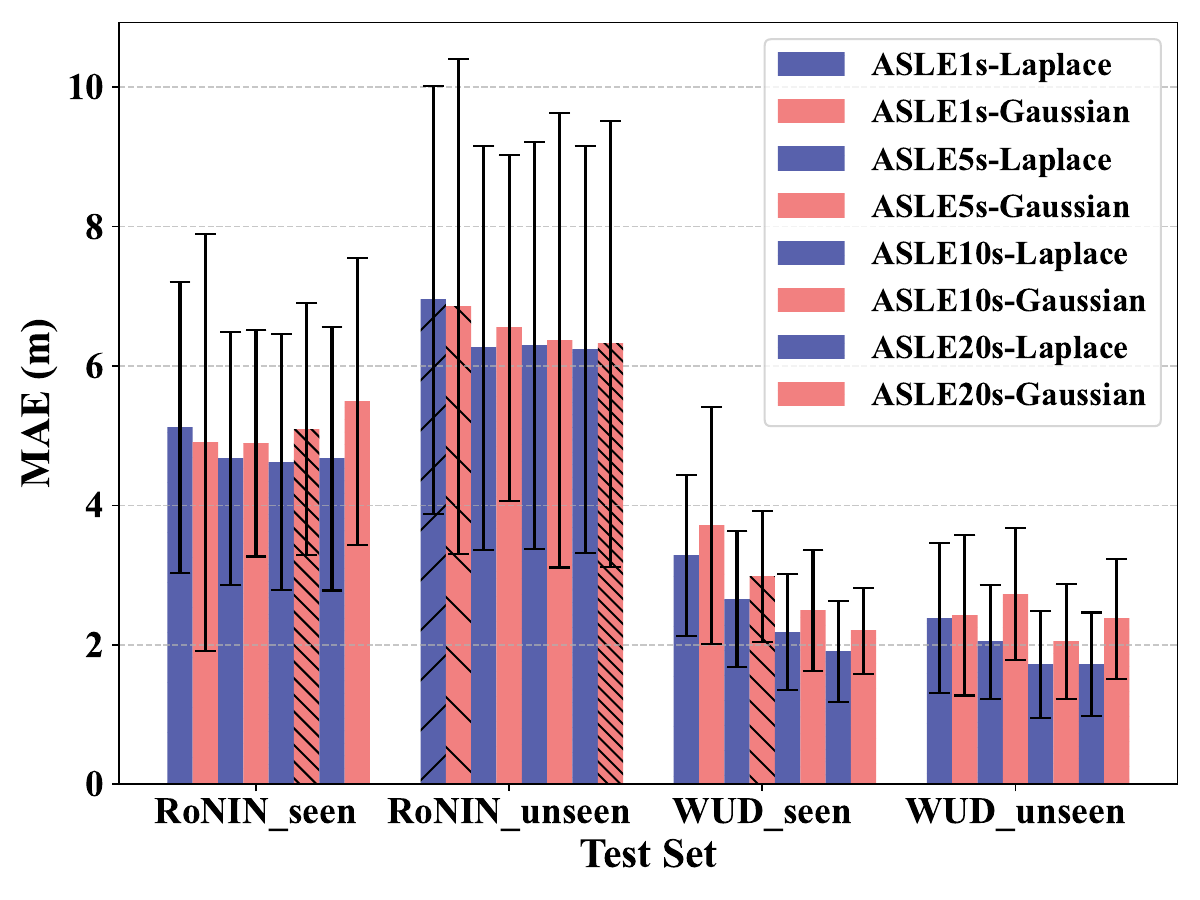}
\Description{MAE comparison when modeling displacement with Gaussian vs. Laplace distributions across different time scales on seen/unseen subjects in RoNIN-ds and WUDataset.}
\caption{Performance of ASLE using Gaussian and Laplace distributions for different time scales across multiple test sets.}
\label{fig_9}
\end{figure}
This experiment aims to compare the effectiveness of using Laplace versus Gaussian distributions to parameterize displacement estimation within the ReNiL framework. Fig.~\ref{fig_9} illustrates the performance of the displacement measured by MAE in seen and unseen subjects from the RoNIN-ds and WUDataset. At short time scales, the performance of both Gaussian and Laplace distributions is comparable, observation errors approximate Gaussian noise when datasets are sufficiently large and sequences are short. However, as the time scale increases, the performance of the Gaussian-based model deteriorates relative to the Laplace-based approach. The Laplace distribution consistently achieves lower MAE, demonstrating its superior capability in capturing the heavier-tailed and more variable displacement distributions characteristic of longer IMU sequences. This advantage of the Laplace distribution can be attributed to its robustness in modeling outliers and abrupt variations in the displacement data, which are less well represented by Gaussian assumptions.
 

\paragraph{\textbf{Consistency and homogeneity analysis of the uncertainty}}
\begin{figure}[!t]
\centering
\includegraphics[width=0.47\textwidth]{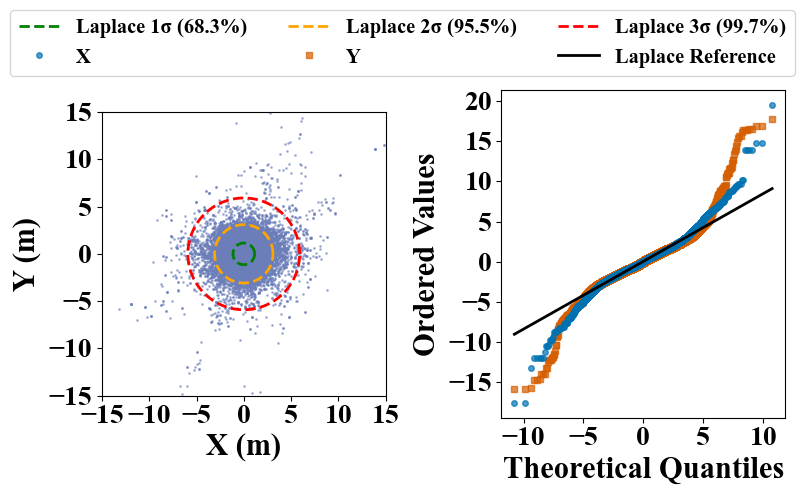}
\Description{Consistency and homogeneity analysis of ASLE uncertainty estimates and real errors. Left: confidence coverage and homogeneity; Right: probability plot showing error compliance.}
\caption{Consistency and homogeneity analysis of the uncertainty estimatios and real errors of ASLE. The left plot shows the consistency and homogeneity, the right plot shows the error compliance.}
\label{fig_18}
\end{figure}
The scale parameter $b$ shares the same units and physical interpretation as the actual displacement error. As shown in Fig.\ref{fig_18}, we performed confidence interval coverage validation and residual normalization tests on the scale-normalized dataset. The left plot in Fig.\ref{fig_18} presents confidence interval coverage rates of 57.7\%, 95.0\%, and 99.2\%, which are slightly below the theoretical confidence levels. This aligns with our previous observation that displacement distributions transition between Gaussian and Laplace models at smaller time scales, corresponding to the relatively conservative $1\sigma$ uncertainty estimates.

The probability plot in the right panel of Fig.~\ref{fig_18} indicates that the model provides reasonable uncertainty estimates for small to medium magnitude errors, while deviations occur for larger errors. This discrepancy is expected since larger errors typically arise over longer time scales, and training data at these scales are comparatively sparse within the dataset.

\paragraph{\textbf{Cross collector, device, and pattern applicability performance}}
\begin{table}[!t]
\caption{Cross-split evaluation under different grouping strategies in WUDataset
(10s window).}
\label{tab:grouping-10s}
\small
\setlength{\tabcolsep}{6pt}
\begin{tabular}{@{}l l l cc cc@{}}
\toprule
\textbf{Grouping strategy} &
\textbf{Training split} &
\textbf{Test split} &
\multicolumn{2}{c}{\textbf{Mean MAE}} &
\multicolumn{2}{c}{\textbf{Variance}} \\
\cmidrule(lr){4-5}\cmidrule(lr){6-7}
& & & \textbf{Train} & \textbf{Test} & \textbf{Train} & \textbf{Test} \\
\midrule
Collectors
& Collectors G1
& Collectors G2
& 1.35 & 1.74 & 0.002 & 0.002 \\

Patterns
& Patterns G1
& Patterns G2
& 1.29 & 1.91 & 0.002 & 0.004 \\

Phones
& Phones G1
& Phones G2
& 1.41 & 1.69 & 0.001 & 0.002 \\

Collectors and patterns
& Collectors G1 + patterns G1
& Collectors G2 + patterns G2
& 1.26 & 1.97 & 0.003 & 0.004 \\

Phones and patterns
& Phones G1 + patterns G1
& Phones G2 + patterns G2
& 1.31 & 1.89 & 0.002 & 0.003 \\

Collectors and phones
& Collectors G1 + phones G1
& Collectors G2 + phones G2
& 1.34 & 1.71 & 0.001 & 0.002 \\
\bottomrule
\end{tabular}
\end{table}
In this section, we employ multiple grouping strategies to conduct experiments. G1 and G2 involve randomly dividing 28 collectors, 9 patterns, and 4 device types into two uniform, disjoint parts according to the grouping strategy. Collectors G1 G2 means dividing the dataset into two disjoint, uniform parts based on collectors, and then using one part for training and the other for testing. Collectors G1 + patterns G1, Collectors G2 + patterns G2 means using the data collected by Collectors G1 with patterns G1 for training, and using the data collected by Collectors G2 with patterns G2 for testing.

In Table~\ref{tab:grouping-10s}. ReNiL demonstrates strong applicability across different collectors, devices, and patterns. It maintains consistent performance, with only moderate variations in mean absolute error (MAE) and variance, even when trained on one set of conditions and tested on another. Specifically, the model shows minimal error across different collectors (1.35 vs. 1.74), devices (1.41 vs. 1.69), and patterns (1.29 vs. 1.91), highlighting its robust cross-collector and cross-device adaptability. Additionally, when combining multiple factors, such as collector and pattern combinations, or device and pattern combinations, the model continues to deliver stable results, with only slight increases in error and variance. This indicates the model's high versatility and robustness in real-world, multi-user, multi-device, and diverse pattern scenarios.

\paragraph{\textbf{Computation overhead}}
\begin{table}[!t]
\centering
\caption{Computation overhead of ReNiL and other methods in RoNIN-ds test set.}
\label{table_6}
\small
\setlength{\tabcolsep}{6pt}
\renewcommand{\arraystretch}{1.2}
\begin{tabular}{lccc}
\toprule
\textbf{Method} & \textbf{Parms} (M) & \textbf{FLOPs (G)} for 20s inference & \textbf{MAE} over RoRes18 \\
\midrule
PDR     & $<0.1$ & $<0.1$ & 513\% \\
IONet   & 1.32   & 5.29   & 326\% \\
RoRes18 & 6.28   & 8.28   & 100\% \\
TLIO    & 6.34   & 8.31   & 98\%  \\
ReNiL   & 2.09  & 0.38   & 91\%  \\
\bottomrule
\end{tabular}
\end{table}
Table \ref{table_6} summarizes the computation overhead for different methods in terms of parameter count, FLOPs, and localization accuracy (represented by MAE relative to RoRes18 in RoNIN-ds). PDR exhibits minimal computational complexity, but it yields significantly worst accuracy. RoRes18 and TLIO have similar parameter counts and computational overhead, while TLIO slightly improves accuracy to 98\% MAE. IONet substantially reduces the size of the model and the computational cost, but achieves only 326\% MAE. ReNiL has 33.3\% the number of parameters and about 4.6\% the FLOPS of RoRes18, and exhibits efficient reasoning capabilities and optimal positioning accuracy. ReNiL improves accuracy while reducing FLOPs by leveraging the fast reasoning capabilities of convolutional networks and MHA's ability to model any scale sequence. 

\paragraph{\textbf{Resource usage comparison}}
\begin{figure}[!t]
\centering
\includegraphics[width=0.60\textwidth]{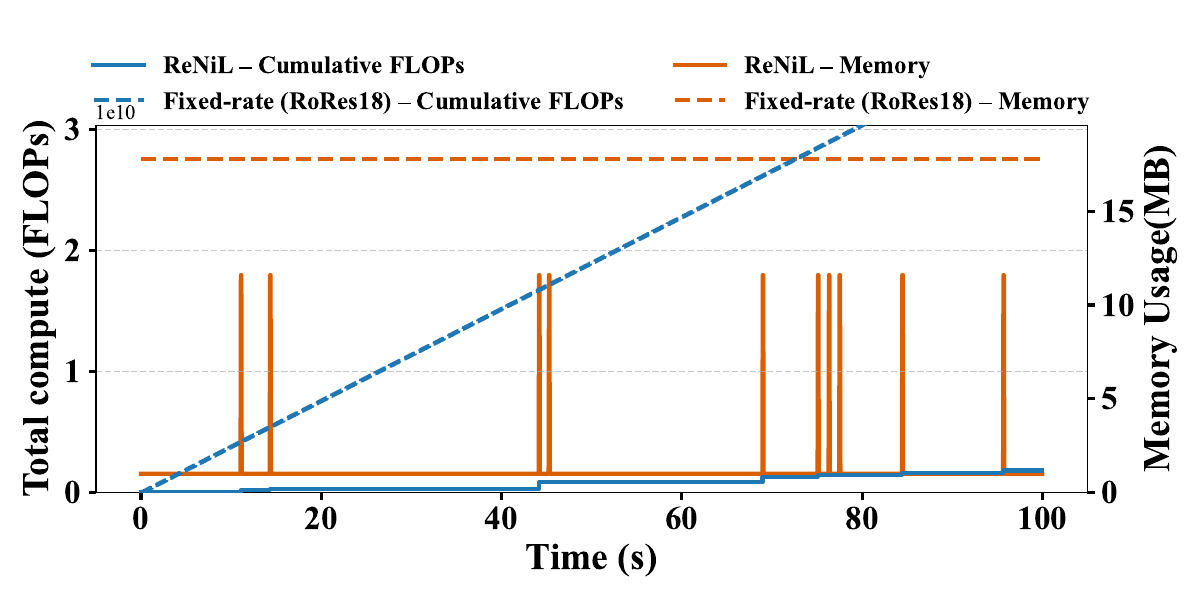}
\Description{Resource usage comparison between ReNiL's ASLE and fixed-rate 5Hz RoRes18 during a 100-second mobile positioning task.}
\caption{An example of a 100-second positioning process running on a mobile phone with 10 positioning requests. Comparison of resource usage between ReNiL's ASLE and fixed-rate 5HZ RoRes18.}
\label{fig_20}
\end{figure}
Fig.\ref{fig_20} shows an example of a 100-second positioning process running on a mobile phone, consisting of 10 positioning requests. ReNiL significantly reduces memory usage, utilizing device memory and computing power almost exclusively at the IPDPs. In contrast, conventional trajectory tracking algorithms maintain high levels of sustained memory usage while linearly increasing FLOPs. Clearly, even in the worst-case scenario, where the application requirement is simply tracking user trajectory, ReNiL still outperforms with a smaller model size and lower single-inference cost.

\paragraph{\textbf{Orientation effectiveness performance}}
\begin{table}[!t]
\centering
\caption{Comparison of device orientation estimation performance, the best result is marked with \best{bold red}.}
\label{table_1}
\small
\setlength{\tabcolsep}{6pt}
\renewcommand{\arraystretch}{1.2}
\begin{tabular}{llrrr}
\toprule
\textbf{Subset} & \textbf{Metric} & \textbf{ReNiL} & \textbf{Mahony} & \textbf{Madgwick} \\
\midrule
\multirow{2}{*}{Vicon} 
  & QAE $\downarrow$ & \best{0.807} & 1.536 & 1.273 \\
  & CS $\uparrow$ & \best{0.733} & 0.350 & 0.538 \\
\midrule
\multirow{2}{*}{Lidar-SLAM} 
  & QAE $\downarrow$ & \best{0.383} & 1.494 & 1.576 \\
  & CS $\uparrow$ & \best{0.837} & 0.429 & 0.329 \\
\bottomrule
\end{tabular}
\end{table}
\begin{figure}[!t]
\centering
\includegraphics[width=0.47\textwidth]{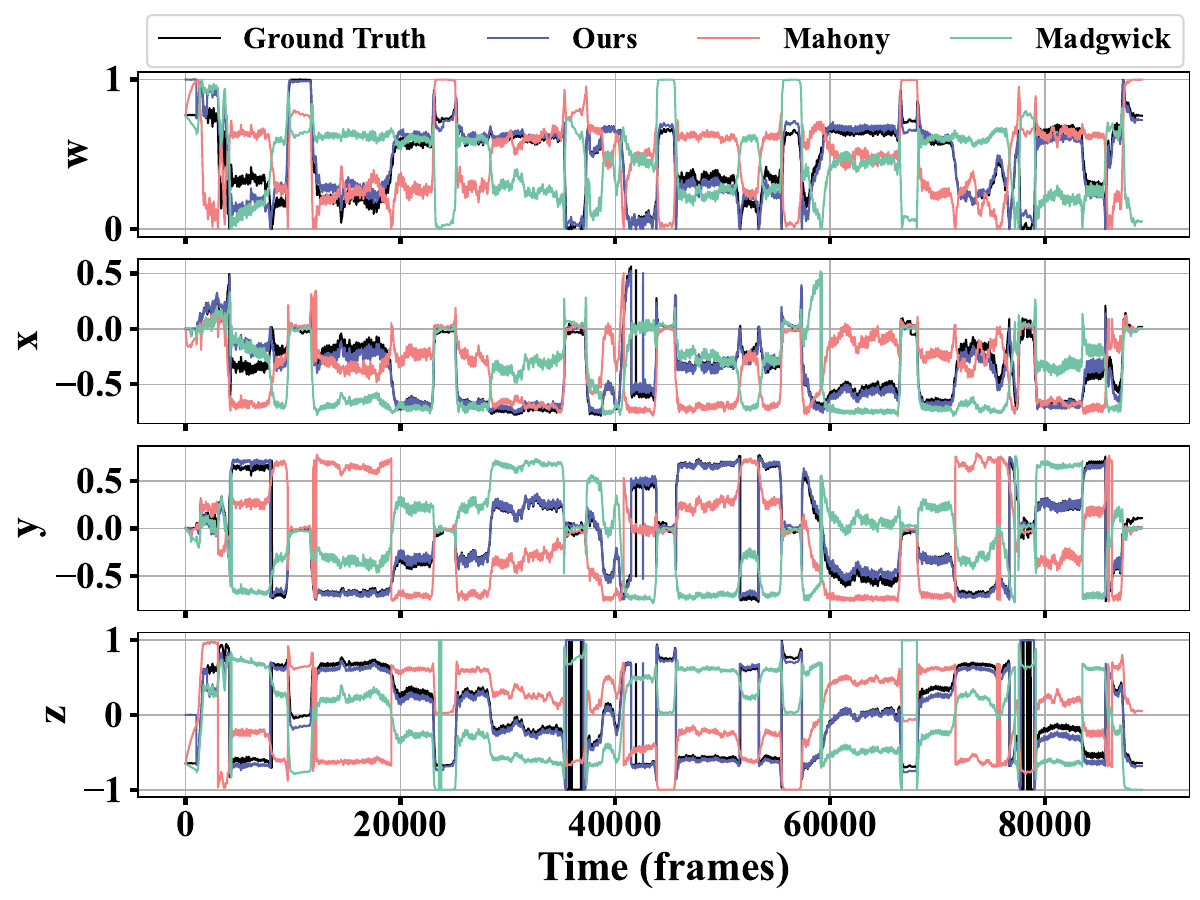}
\Description{Comparison of device orientation estimation for ReNiL, Mahony, and Madgwick methods against ground truth.}
\caption{Comparison of device orientation estimation results for different methods and the ground truth}
\label{fig_5}
\end{figure}
Fig.~\ref{fig_5} shows the results of these methods and the ground truth for device orientation during one example. As shown in Table~\ref{table_1}, our method significantly outperforms Mahony and Madgwick in estimating the orientation of the device during pedestrian movement, both when using Vicon and Lidar-SLAM systems. Our method converges the device orientation estimation framework to the navigation coordinate system in a short time, and maintains stability and effectiveness even under complex pedestrian motion patterns and magnetic field changes. In contrast, Mahony and Madgwick tend to experience orientation estimation fluctuations and heading drift due to the complexity of pedestrian motion patterns and magnetic field disturbances.

\subsection{Application Performance in Real World}

We demonstrate the high usability of our approach by integrating it with another behavioral semantics method that relies exclusively on inertial sensors. The experiment was conducted in a standard office environment, covering a plane area of approximately 100 meters in the east-west direction and 80 meters in the north-south direction. The inbound data was collected within a $14m \times 18m$ office space, while the outbound data was collected in the remaining public areas. In total, approximately 1.5 hours of activity data were collected, with ground truth obtained using Lidar-SLAM.

\paragraph{\textbf{Behavioral semantic map construction}}
\begin{figure*}[!t]
\centering
\includegraphics[width=0.95\textwidth]{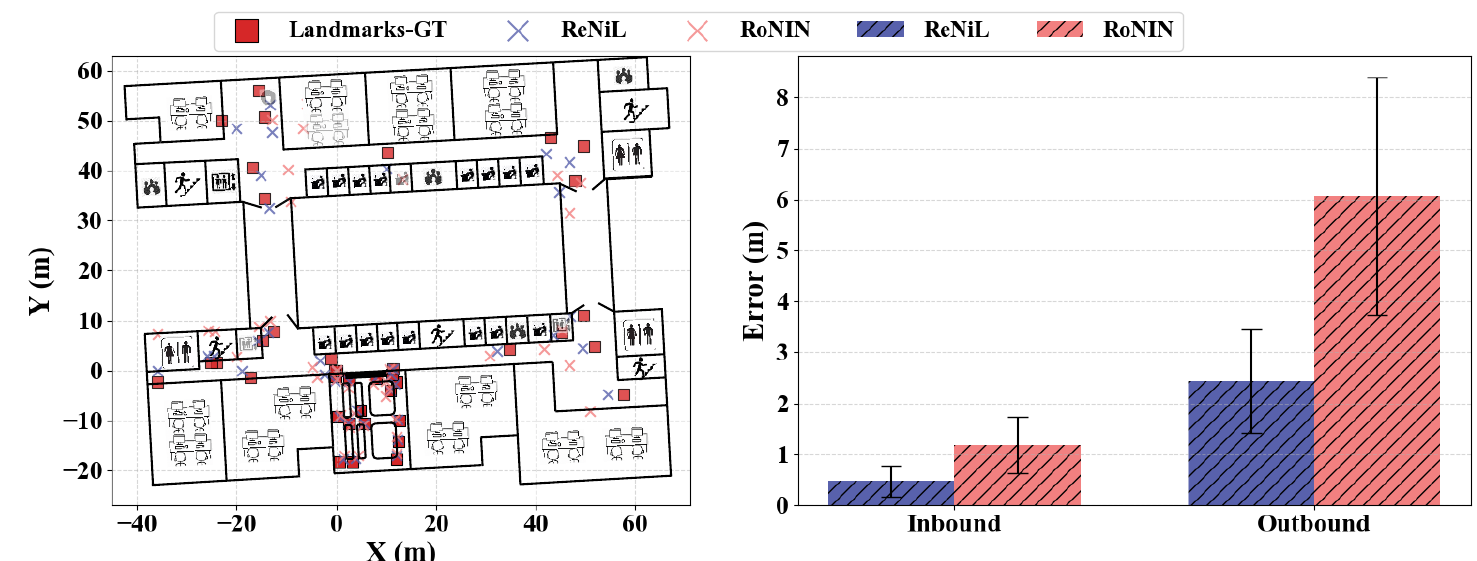}
\Description{Behavioral semantic map construction comparison between ReNiL, RoNIN, and ground truth, including quantitative landmark error comparison.}
\caption{Behavioral semantic map construction results between ReNiL, RoNIN and ground truth. Quantitative comparison of the semantic landmarks construction error between ReNiL and RoNIN.}
\label{fig_13}
\end{figure*}

We implemented the construction of behavioral semantic landmarks based on the approach introduced by NIU~\cite{niu2020atlas,10.1145/3749455}, with two key modifications: instead of using the step-by-step window integration method for displacement estimation, we employed our IPDPs estimation framework. We incorporated the motion-aware orientation filter to ensure that the constructed map's coordinate system aligns with the ENU reference frame.

The construction result for the behavioral semantic map is shown in the left plot of Fig.~\ref{fig_13}. Our method demonstrates a clear advantage in estimating the true position of the semantic landmarks compared to the RoRes18 construction method. As indicated in the figure, our approach significantly outperforms RoRes18 in both inbound and outbound areas, with estimated landmarks located closer to the ground truth. It is a reasonable result that the outbound performance exhibits slightly higher error compared to the inbound, because of the challenges of human motion complexity and data sparsity in the outbound environment. For a more detailed quantitative comparison of the semantic landmark construction error, please refer to the right part of Fig.~\ref{fig_13}. The error bars show that our method achieves less than half the absolute error compared to RoRes18. 

\paragraph{\textbf{Pure IPDPs task}}
\begin{figure}[!t]
\centering
\includegraphics[width=0.47\textwidth]{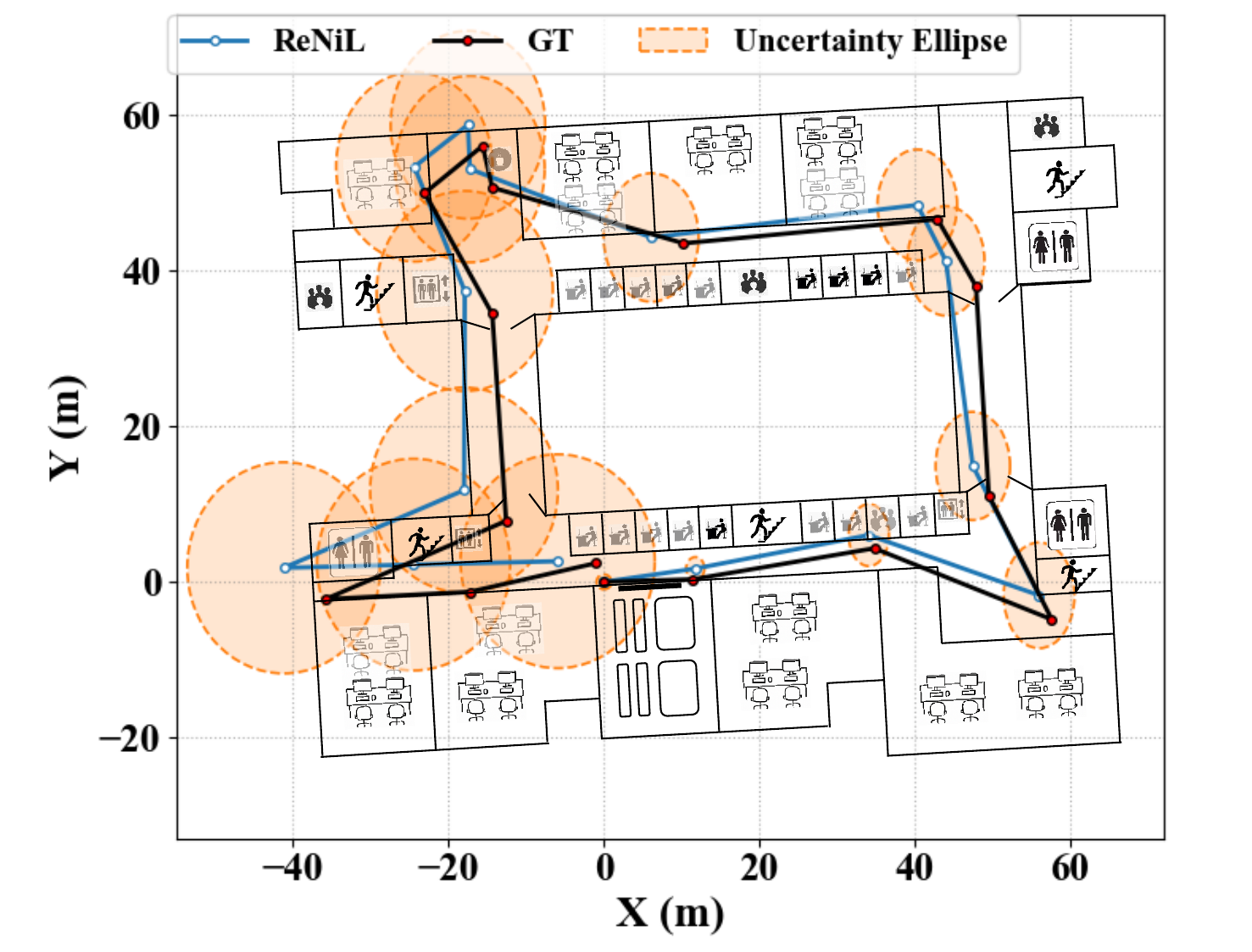}
\Description{Visualization of pure IPDP task showing displacement estimates and 99.7\% uncertainty ellipses.}
\caption{Pure IPDPs task. The orange dashed circles represent 99.7\% uncertainty ellipses.}
\label{fig_15}
\end{figure}

We demonstrated an example of an pure IPDP task. ASLE uses past historical information as a priori and builds a complete Bayesian inference chain through the Gaussian-exponential mixture. As shown in Fig.\ref{fig_15}, the inference process shows excellent and robust displacement estimation performance and error estimation capabilities. For long-running inference processes, the ground truth can always be guaranteed to be within the given error ellipse. This illustrates the effectiveness of using the Gaussian-exponential mixture to conditionally Gaussianize the Laplace distribution.

\paragraph{\textbf{Fusion of behavioral semantics and ASLE}}
\begin{figure}[!t]
\centering
\includegraphics[width=0.47\textwidth]{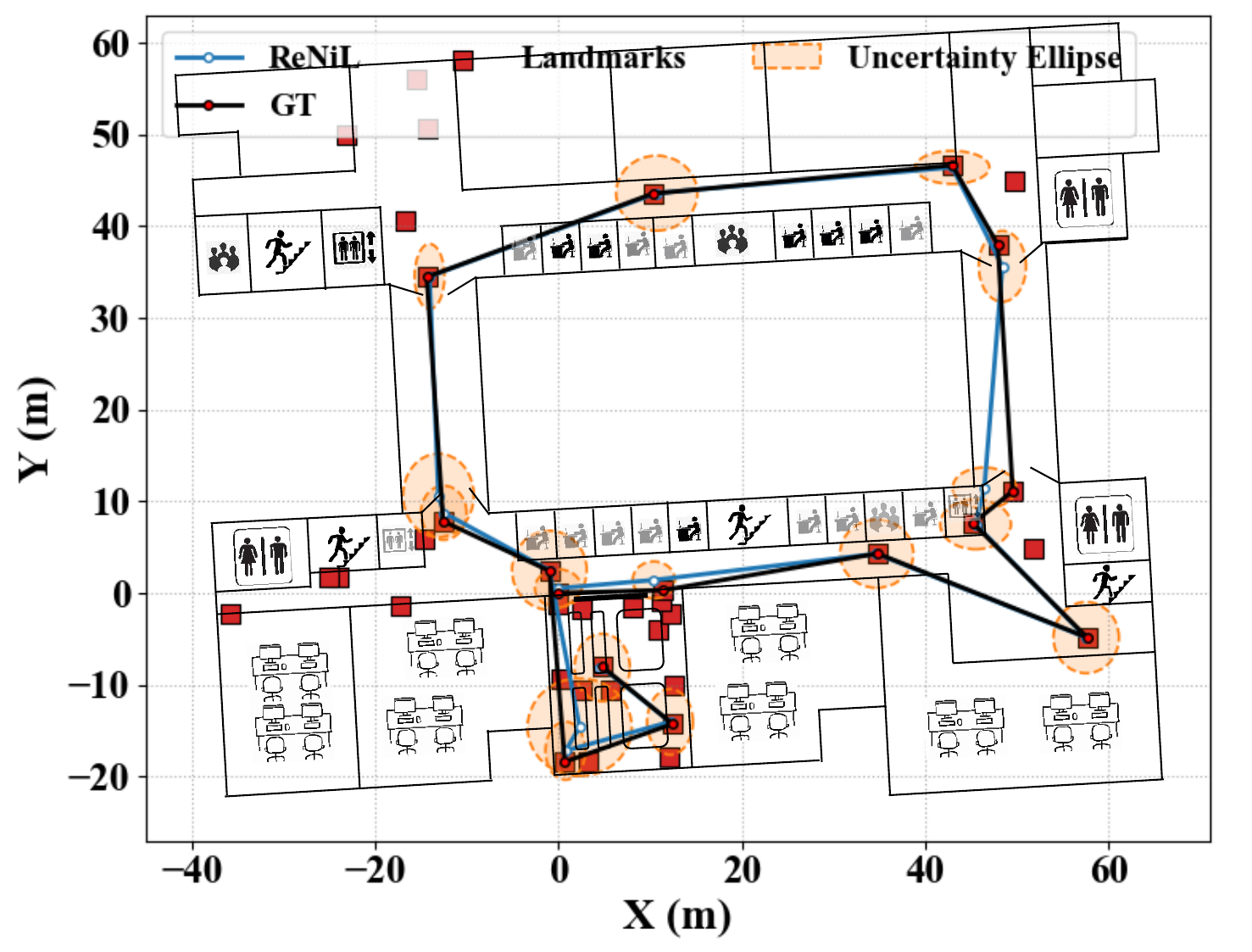}
\Description{Fusion process of behavioral semantics and ASLE showing 99.7\% uncertainty ellipses and alignment with ground truth.}
\caption{Fusion of behavioral semantics and ASLE in the ReNiL framework. The orange dashed circles represent 99.7\% uncertainty ellipses.}
\label{fig_14}
\end{figure}

We demonstrate an example of Rao-Blackwellised Kalman Gibbs filter for Bayesian process within the ReNiL framework. As shown in Fig.~\ref{fig_14}, the process begins when a pedestrian triggers a behavior landmark in a behavioral semantic library. The ReNiL framework uses IMU data collected between time-adjacent semantic landmarks as input. In this system, the ASLE is loosely coupled with the observed behavior landmark, and its associated observation noise covariance matrix is incorporated into the fusion process. 

Fig.~\ref{fig_14} visually demonstrates the effectiveness of our method in integrating IMU data with behavioral semantics for accurate location estimation. The Kalman filter helps refine the position estimates by merging the IMU data with observed behavior landmarks. This approach accounts for the increased uncertainty in some areas while still keeping the predicted location aligned with the ground truth. It can be observed that the true position of IPDPs is always within the estimated uncertainty ellipse. The close alignment between the predicted location and the ground truth shows the robustness of the fusion process. It also highlights the effectiveness of combining behavioral semantic landmarks with IMU-based positioning.

\section{Conclusion}
\label{sec:conclusion}

This paper introduces ReNiL, an event-driven Bayesian inertial-localization framework that shifts the focus from dense trajectory tracking to accurate position and uncertainty estimation at IPDPs. ReNiL combines the motion-aware orientation filter with ASLE, enabling robust displacement and homogeneous uncertainty regression for IMU sequences of any scale. ReNiL also provides a general Bayesian process that accommodates standalone inertial use or loose coupling with external sensors. Future work will explore (i) cross-device and cross-activity generalization without domain-specific fine-tuning, (ii) real-time deployment on ultra-low-power edge processors, and (iii) tighter fusion with GNSS or other signals to further improve availability in challenging environments. We believe ReNiL and the IPDPs perspective establish a solid foundation for next-generation pedestrian localization services that are accurate, efficient, and uncertainty-aware.

\bibliography{ref}
\end{document}